\documentclass{optica-article}
\journal{opticajournal} % for journals or Optica Open
\articletype{Research Article}

\usepackage[utf8]{inputenc}

\usepackage{calc}
\usepackage{tikz}
\usetikzlibrary{calc}
\usetikzlibrary{shapes.geometric}
\usepackage{pgfplots}
\pgfplotsset{compat=1.18}
\newlength{\tfwidth}
\newlength{\tfheight}
\newlength{\tfxa}
\newlength{\tfxb}
\newlength{\tfya}
\newlength{\tfyb}
\newcommand{\trimFigWithBox}[6]{%
\setlength\fboxsep{0pt}%
\setlength\fboxrule{1.0pt}% border thickness
\fbox{\includegraphics[width=#2, clip, trim=#3 #4 #5 #6]{#1}}%
}
\newcommand{\trimFigNoBox}[6]{%
\setlength\fboxsep{1pt}% note: make this 1pt and rule thickness zero so box size matches that below
\setlength\fboxrule{0.0pt}% border thickness
\fbox{\includegraphics[width=#2, clip, trim=#3 #4 #5 #6]{#1}}%
}
\newcommand{\trimFigHeightWithBox}[6]{%
\setlength\fboxsep{0pt}%
\setlength\fboxrule{1.0pt}% border thickness
\fbox{\includegraphics[height=#2, clip, trim=#3 #4 #5 #6]{#1}}%
}
\newcommand{\trimFigHeightNoBox}[6]{%
\setlength\fboxsep{1pt}% note: make this 1pt and rule thickness zero so box size matches that below
\setlength\fboxrule{0.0pt}% border thickness
\fbox{\includegraphics[height=#2, clip, trim=#3 #4 #5 #6]{#1}}%
}
\newsavebox\figBox

\newcommand{\trimw}[6]{%
\sbox\figBox{\includegraphics{#1}}
\setlength{\tfwidth}{\the\wd\figBox}
\setlength{\tfheight}{\the\ht\figBox}
\setlength{\tfxa}{\tfwidth*\real{#3}}%
\setlength{\tfxb}{\tfwidth*\real{#4}}%
\setlength{\tfya}{\tfheight*\real{#5}}%
\setlength{\tfyb}{\tfheight*\real{#6}}%
\trimFigNoBox{#1}{#2}{\tfxa}{\tfya}{\tfxb}{\tfyb}%
}

\newcommand{\trimwb}[6]{%
\sbox\figBox{\includegraphics{#1}}
\setlength{\tfwidth}{\the\wd\figBox}
\setlength{\tfheight}{\the\ht\figBox}
\setlength{\tfxa}{\tfwidth*\real{#3}}%
\setlength{\tfxb}{\tfwidth*\real{#4}}%
\setlength{\tfya}{\tfheight*\real{#5}}%
\setlength{\tfyb}{\tfheight*\real{#6}}%
\trimFigWithBox{#1}{#2}{\tfxa}{\tfya}{\tfxb}{\tfyb}%
}

\newcommand{\trimh}[6]{%
\sbox\figBox{\includegraphics{#1}}
\setlength{\tfwidth}{\the\wd\figBox}
\setlength{\tfheight}{\the\ht\figBox}
\setlength{\tfxa}{\tfwidth*\real{#3}}%
\setlength{\tfxb}{\tfwidth*\real{#4}}%
\setlength{\tfya}{\tfheight*\real{#5}}%
\setlength{\tfyb}{\tfheight*\real{#6}}%
\trimFigHeightNoBox{#1}{#2}{\tfxa}{\tfya}{\tfxb}{\tfyb}%
}

\newcommand{\trimhb}[6]{%
\sbox\figBox{\includegraphics{#1}}
\setlength{\tfwidth}{\the\wd\figBox}
\setlength{\tfheight}{\the\ht\figBox}
\setlength{\tfxa}{\tfwidth*\real{#3}}%
\setlength{\tfxb}{\tfwidth*\real{#4}}%
\setlength{\tfya}{\tfheight*\real{#5}}%
\setlength{\tfyb}{\tfheight*\real{#6}}%
\trimFigHeightWithBox{#1}{#2}{\tfxa}{\tfya}{\tfxb}{\tfyb}%
}
\usepackage{geometry}
\usepackage{subcaption} % subfigures
\usepackage{adjustbox}  % scale tables to fit on the page
\usepackage{amsmath}
\usepackage{amsfonts}
\usepackage{xcolor}
\usepackage{bm}
\usepackage{boldline}
\usepackage{soul}
\usepackage{multirow}

\newcommand{\noise}{{n}}
\newcommand{\direct}{{d}}

\newcommand{\abel}{\mathcal{A}}

\newcommand{\density}{\rho}
\newcommand{\m}{{m}}

\usepackage[normalem]{ulem}

\newcommand{\das}[1]{{\color{red}#1}}

\date{Aug. 2023}
\newcommand{\remove}[1]{}

\begin{document}

\title{Reconstructing Richtmyer–Meshkov instabilities from noisy radiographs using low dimensional features and attention-based neural networks}

\author{Daniel A. Serino,\authormark{1,*} 
Marc L. Klasky,\authormark{1} 
Balasubramanya T. Nadiga,\authormark{1}
Xiaojian Xu,\authormark{2}
and Trevor Wilcox\authormark{1}
}

\address{\authormark{1}Los Alamos National Laboratory, Los Alamos, NM 87545, USA\\
\authormark{2}Dept. of Electrical and Computer Engineering, University of Michigan, Ann Arbor, MI 48109
}

\email{\authormark{*}dserino@lanl.gov} %% email address is required; see note below about the corresponding author designation

\begin{abstract} 
A trained attention-based transformer network can robustly recover the complex topologies given by the Richtmyer-Meshkoff instability from a sequence of hydrodynamic features derived from radiographic images corrupted with blur, scatter, and noise. This approach is demonstrated on ICF-like double shell hydrodynamic simulations.
The key component of this network is a transformer encoder
that acts on a sequence of features extracted from noisy radiographs.
This encoder includes numerous self-attention layers
that act to learn temporal dependencies in the input sequences and increase the expressiveness of the model.
This approach is demonstrated to 
exhibit an excellent ability to 
accurately recover the Richtmyer-Meshkov instability
growth rates, 
even despite the gas-metal interface
being greatly obscured by radiographic noise.
\end{abstract}

%\maketitle

\section{Introduction}

The ability to isolate precise material interfaces from radiographic images is of paramount importance in describing a broad array of physical phenomena including a number of hydrodynamic instabilities including the Richtmyer Meshkoff instability (RMI)~\cite{richtmyer1954taylor,meshkov1969instability}, the Rayleigh Taylor 
%(RT) 
instability~\cite{rayleigh1882investigation,taylor1950instability,sharp1984overview}, and the Kelvin–Helmholtz 
%(KH) 
instability~\cite{drazin1970kelvin}. These phenomena arise in a broad range of situations from astronomical size events like supernova collapse \cite{kane1999evaluation}, to microscopic events such as gas bubble sonoluminescence \cite{holzfuss2008surface} and the supersonic combustion of ramjets~\cite{goodwin2018premixed}.  Indeed, extraction of the peaks and troughs associated with these phenomena at the material interfaces, RMI in particular, is essential in capturing the growth rates and optimizing designs to mitigate these instabilities. Here we focus on the RMI that originates from the interaction of a shock wave with an interface separating two materials with different densities.  In most cases an initial perturbation will be amplified following the passage of the shock.  A deposition of baroclinic vorticity that is practically-instantaneous and results from a misalignment between the pressure gradient across the shock and the local density gradient across the interface leads to a subsequent (slower) growth of the perturbation.    

While extensive examinations and reviews of the RMI
have been performed over the past sixty years,
most of of the experimental work to validate the RMI has been performed within shock tubes
using rectangular geometry with gaseous or fluid materials,
although some recent work has been performed in cylindrical geometry%
~\cite{rupert1992shock,brouillette2002richtmyer,zhou2021rayleigh,leinov2009experimental,holmes1999richtmyer,zhou2017rayleigh,zhang1998numerical}.
Additionally, some recent works have also examined the RMI in solid-solid settings
to estimate the strength in ductile materials~\cite{prime2019tantalum,prime2017estimation}.

\begin{figure}%[htbp]
  \centering
  \resizebox{.35\columnwidth}{!}{%
  \begin{tikzpicture}
  \useasboundingbox (0,0.25) rectangle (15,22);  

   \coordinate (A) at (7.5, 1);
   \coordinate (B) at (3, .5);

\node[isosceles triangle,
    isosceles triangle apex angle=30,
    draw,
    rotate=270,
    fill=blue!60,
    minimum size=11cm, 
    anchor=apex] (Alum) at (A){};
\node[isosceles triangle,
    isosceles triangle apex angle=30,
    draw,
    rotate=270,
    fill={rgb,1:red,.85; green,.85; blue,.85},
    minimum size = 9.35cm, 
    anchor=apex] (Foam) at (A){};  
\node[isosceles triangle,
    isosceles triangle apex angle=30,
    draw,
    rotate=270,
    fill={rgb,1:red,.7; green,.7; blue,.7},
    minimum size = 2.95cm, 
    anchor=apex] (Be) at (A){};  
\node[isosceles triangle,
    isosceles triangle apex angle=30,
    draw,
    rotate=270,
    fill={rgb,1:red,.2; green,.2; blue,.2},
    minimum size = 2.55cm, 
    anchor=apex] (W) at (A){};     
\node[isosceles triangle,
    isosceles triangle apex angle=30,
    draw,
    rotate=270,
    fill={rgb:orange,1;yellow,2;pink,1},
    minimum size=2.15cm, 
    anchor=apex] (DT) at (A){};  

\node[scale=2.5, anchor=west, align=center] at (13, 21.5) {1100 $\mu$m};
\node[scale=2.5, anchor=west, align=center] at (12.5, 18.5) {982 $\mu$m};
\node[scale=2.5, anchor=west, align=center] at (9.5, 7) {290 $\mu$m};
\node[scale=2.5, anchor=west, align=center] at (9, 5.75) {260 $\mu$m};
\node[scale=2.5, anchor=west, align=center] at (8.5, 4.5) {230 $\mu$m};

\node[scale=2.5, align=center] at (-1, 20) {Aluminum \\ 2.7 g/cc};
\node[scale=2.5, align=center] at (-1, 13) {Low Density Foam \\ 35 mg/cc};
\node[scale=2.5, align=center] (a) at (-1, 7) {Be Tamper \\ 1.85 g/cc};
\node[scale=2.5, align=center] (b) at (-1, 4.75) {W Pusher \\ 19.41 g/cc};
\node[scale=2.5, align=center] (c) at (-1, 2.5) {DT \\ 0.2 g/cc};

\draw[->, line width=5] (a) -- ($(Be) + (-1.75, .75)$);
\draw[->, line width=5] (b) -- ($(W) + (-1.5, .5)$);
\draw[->, line width=5] (c) -- ($(DT) + (-1.0, -1)$);

  \end{tikzpicture}  
  }
  \caption{Example double shell capsule specification based on the 1.06 MJ yield design from Ref.~\cite{merritt19}.
  }
  \label{fig:icf}
  %\vspace{-0.1in}
\end{figure}
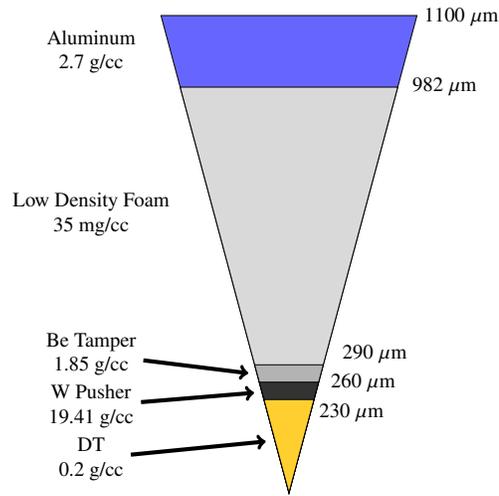

The emergence of Inertial Confinement Fusion (ICF) as a potential power source has been a major impetus for the continued examination of the RMI~\cite{zhou2021rayleigh,brouillette2002richtmyer}.  
However, the spherical convergent geometry as well as large attenuation present in these experiments precludes previous analysis methods and diagnostics.
For example, planar laser induced fluorescence, planar Rayleigh scattering, and particle image velocimetry. Consequently, examination of the RMI instability in promising ICF concepts, such as the double shell capsules, depicted in Figure~\ref{fig:icf} demand the development of new experimental and analysis techniques to capture details of the RMI whose behavior and control are thought to be crucial for continued progress in ICF~\cite{brouillette2002richtmyer,prestridge2000validation}.
That is, the RMI may induce mixing due on the interface between
the outer shell and the inner fuel and in so doing inhibit the fusion reaction
and can consequently be a limiting factor in the energy produced~\cite{emery1991hydrodynamic,ishizaki1997propagation,lindl1992progress,taylor1997saturation}.
 %Finally, emerging second generation ICF design concepts such as double shells further necessitate the need to obtain additional understanding of the role of instabilities in systems in which a heavy metal interfaces with an interior gas.

\begin{figure}[tb]
  \centering
  \includegraphics[trim=4cm 3cm 2.7cm 2cm, clip,width=.99\textwidth]{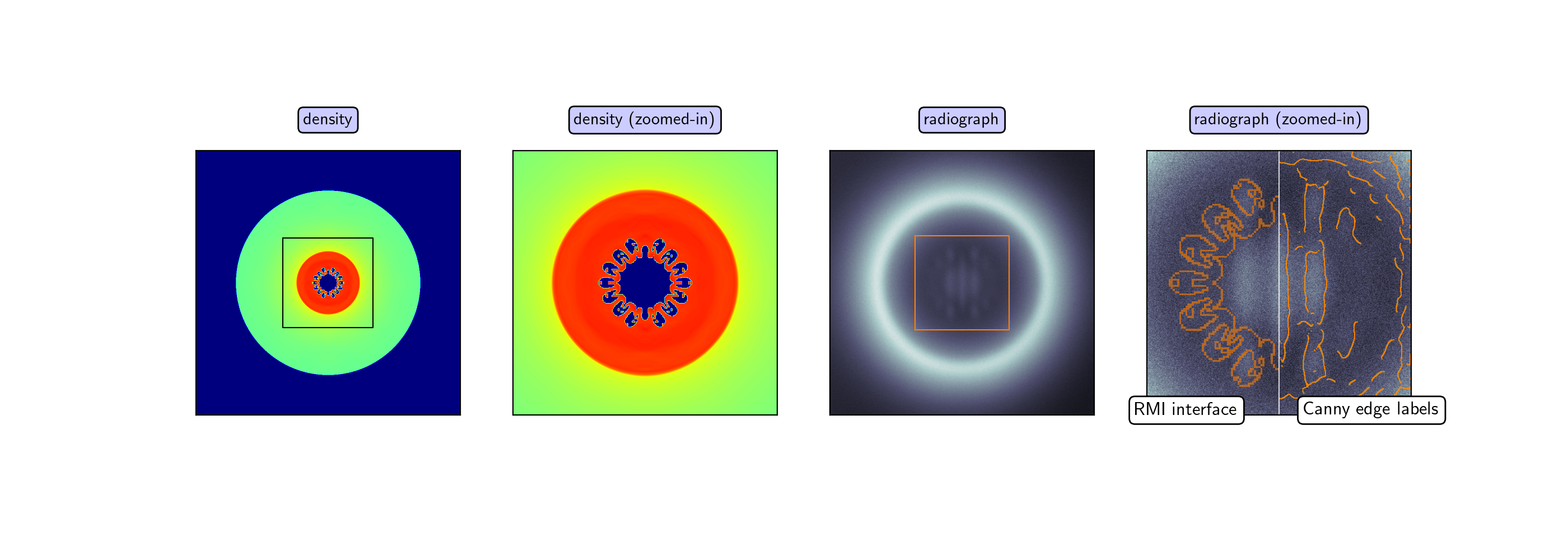}
  \caption{Sample $(r, z)$ projection of the density (1$^{\rm st}$ column), 
  zoomed-in view of the Richtmyer–Meshkov interface (2$^{\rm nd}$ column),
  synthetic radiograph (3$^{\rm rd}$ columns), 
  and a zoomed-in view of the radiograph (4$^{\rm th}$ columns)
  labeled with the RMI interface (left half) 
  and Canny edge labels (right half).
  }
  \label{fig:sampledensitysingle}
\end{figure}  

As an illustration in the difficulty in locating the interfaces necessary to quantitatively determine the RMI growth rates in spherical geometries, we present the result of a ICF hydrodynamic simulation in which a shock wave generated during the implosion of a 
nearly-spherical double shell capsule into a gas medium 
%(with a small perturbation on the interior surface) 
interacts with the irregular surface at the gas-metal interface during rebound to generate an RMI.  It should be noted that this configuration is of particular interest in the proposed double shell ICF simulations as illustrated in Figure~\ref{fig:icf}.

While the material interface in the hydrodynamic simulation is clearly observable
and easily captured by applying traditional image processing methods,
the observation from the noisy radiographic image makes the identification of the peaks and troughs
(which are necessary to quantitatively characterize the growth rates of the RMI) problematic.
Indeed, as may be observed from Figure~\ref{fig:sampledensitysingle},
traditional edge detection algorithms are generally insufficient to accurately find these features.
Furthermore, these methods (Sobel, Canny, and Laplacian of Gaussian edge filters)
generally require fine-tuning of hyper-parameters associated with the method,
which may lead to significant uncertainty/error in the results.
Finally, the sensitivity of the RMI to the initial characteristics of the perturbed surface
further complicate the quantitative determination of growth rates.

This paper reports on the development of a new physics-based density reconstruction method
capable of capturing, 
through a 
%density reconstruction and 
subsequent interface finding algorithm,
RMI growth rates of ICF double shell implosions in a spherically convergent geometry using
low dimensional features that are robustly identifiable from dynamic radiographic images.
This approach may enable experimental validation in a regime that, to our knowledge, been largely unexplored.
% brief paper outline here?  the next section still seems like "intro" to me, btw.
% perhaps make it a subsection of intro?

\section{Density Reconstruction via Dynamic Radiography}

To examine the dynamic evolution of materials undergoing strong deformation
in material science, shock physics, and ICF,
dynamic radiography is routinely used as an experimental diagnostic%
~\cite{lifshin2008x, swift2018absolute, rygg20142d,hossain2022high}.
Indeed, capturing the dynamic evolution of the material interface is essential in quantifying the growth rates of the RMI.  As previously discussed, the limitations of traditional image processing algorithms require consideration of alternative techniques for capturing this interface with sufficient accuracy to infer the growth rates from noisy radiographs.
One such technique for capturing the material interface is to use traditional density reconstruction techniques followed by applying gradient-based approaches
to locate the material interfaces associated with the RMI growth rates.
That is, an inversion of the projection data may first be performed to obtain the density field.
Traditional methods of performing density inversions date back to Radon~\cite{radon1917uber}.
Over the past several decades, image reconstruction methods have evolved from simple analytical methods
such filtered-back projection (FBP) methods, X-ray CT (e.g. Feldkamp-Davis-Kress or FDK methods),
and the Inverse Abel Transform for axisymmetric systems \cite{cormack1963representation,feldkamp1984practical,bracewell1986fourier}
to models that enable inclusion of poly-energetic sources and also attempt to address the noise field (e.g. statistical noise as well as scatter).  The models are complex non-linear, non-convex forward models and employ iterative reconstruction techniques~\cite{elbakri2002statistical}.  Iterative reconstruction (IR) algorithms are based on more sophisticated models for the imaging system's physics and models for sensor and noise statistics.  These methods are often called model based image reconstruction (MBIR) methods or statistical image reconstruction (SIR) methods \cite{ravishankar2019image}.  
Generally, density reconstruction algorithms using dynamic experimental radiographic data with complex noise fields (e.g., non-Gaussian noise, scattered radiation, complex beam dynamics, etc.) still encounter difficulty when sufficient accuracy is required to extract the intricate details necessary to characterize the RMI phenomena~\cite{hossain2022high}.  Furthermore, the nature of these methods introduces a certain degree of regularization that may impact the ability to accurately preserve edges.  Research continues in this area to improve the ability to preserve edges in settings where uncertainty in the characteristics of scatter and noise are present~\cite{pasha2023computational,yu2002edge,delaney1998globally}.
%This difficulty arises due to the inability to exactly represent various aspects of the radiographic measurement system, such as scatter, beam spot movement, beam-target interactions, beam dynamics repeatability, and aspects of the image formation process. 

More recently, machine learning (ML) approaches have been applied to radiographic reconstruction. Many of these ML architectures have outperformed IR methods by a large margin at a specified degradation level \cite{unberath2019enabling,sajed2023effectiveness}.
Similar to traditional non-ML-based algorithms, ML-based approaches experience difficulty with complex noise fields.
ML architectures that are trained using a specific noise distribution typically fail to obtain favorable results in the presence of out-of-sample noise~\cite{sun2021degradation,huang2022physics}.  
This problem can be partially addressed by training a number of models targeting each level of degradation. 
However, in many cases, the noise depends on the object and other imaging artifacts and therefore it is impractical to produce a general noise model to cover all unknown eventualities. 
In recent work~\cite{huang2022physics}, a machine learning-based Wasserstein generative adversarial network (WGAN) was developed to reconstruct density fields from noisy radiographic projections. 
This approach demonstrated excellent performance in reconstructing density
for testing cases that had the same noise properties as the training set.   
However, when testing noise levels outside of the training population,
the reconstructions exhibited a rapid degradation in performance.  
Interestingly, degradation in performance was observed
even when the noise level was reduced from the level used in the training.

In light of the inability to train a general denoising algorithm to address the issues associated
with out-of-population noise and other imaging artifacts,
we instead use a class of new ML-based methods that rely on features
that are robustly identifiable from a series of dynamic radiographs.
These methods are composed of two major components;
one component is responsible for feature extraction, in this case the robust out-going shock, 
and another component performs density reconstructions from the extracted shock. 
This approach has been previously demonstrated on a relatively simple 1D system,
where robust features were combined with the underlying hydrodynamic equations of motion
to produce density reconstructions \cite{hossain2022high}.
This methodology outperformed a traditional, direct radiograph-to-density reconstruction method
in the presence of scatter.
Furthermore, the method was also capable of generating families of solutions consistent with the observed features and a methodology for examining the uncertainty in the predictions was proposed. % ref

To investigate the RMI we consider a test problem of shock propagation due to the 
implosion of a gas-filled ICF double shell capsule with a sinusoidal perturbation on the interior surface.
Numerous simulations were performed with variable initial conditions and material parameters 
to produce a data set of density field sequences in time. 
Synthetic radiographs were produced for each time snapshot of density and a noise field was generated to mimic a realistic experimental setup. 
Ground truth features were generated using the density fields
corresponding to coefficients of a curve fit for the shocks.
We focused on a features-to-density network, 
for which we investigated two approaches 
including a generative variational autoencoder (VAE) 
approach called the ShockDecoderViT based
on the vision transformer~\cite{dosovitskiy2021an},
and a deterministic structure-preserving model 
called the Mass-Conserving Transformer,
based on the original transformer~\cite{Vaswani17} to reconstruct the density fields
to enable the subsequent extraction of the gas/metal interface
and
%, as described in Section XX, to 
%quantitatively determine 
the RMI growth rates.
Each architecture transforms a temporal sequence 
of features into their corresponding density fields using networks that incorporate attention layers.
In previous work,
attention blocks increased the expressive power of the networks
by correlating temporal dependencies in the sequences \cite{Vaswani17}. 
In this paper, we demonstrate that incorporating attention in the features-to-density networks
improves density reconstruction errors relative to a similar network without attention.
The density fields are then processed to obtain the gas/metal interface
to obtain the RMI growth rates.

To test the effectiveness of the features-to-density network in the full radiograph-to-density pipeline,
we developed a noise model to characterize the errors of a feature extractor network.
For the noise model,
we used the results from a feature extractor network developed in~\cite{xiaojian}
that consists of a 
convolutional neural network (CNN) with an image Fourier feature encoding (IFFE) layer.
This network was trained on the noisy radiograph-to-features problem
and is capable of recovering the shock features with sub-pixel accuracy.
We trained the features-to-density network 
using noiseless features 
and applied the noise model during testing.
We show that the features-to-density network
remains robust 
in the presence of significant amounts 
of out-of-sample noise.
Using the results of these ML methods,
we investigated the limits of the radiograph-to-density pipeline
in reconstructing the much higher dimensional phenomena present. 
Using the density fields we then assessed the ability of the ML method
to reconstruct the gas metal interface attributed to the RMI with lower dimensional features,
such as the outgoing shock captured from the radiographs. 
This pipeline provides a possible means to address the difficulty in validating simulations of 3D ICF double shell RMI experiments.

In the remainder of this work we present details of our new approach for performing density
reconstructions and subsequent extraction of the gas-metal interface necessary for characterizing RMI phenomena using the robust features extracted from the dynamic radiographic images.
Section~\ref{sec:ProblemSetup}
introduces the dynamic radiography problem and model formulation.
Section~\ref{sec:DataGeneration}
introduces details of the hydrodynamic ICF test problem and data generation.
Section~\ref{sec:ArchitectureDescription}
presents two machine learning approaches for determining the density fields
from robust features identified from a dynamic radiographic sequence. 
%from the robust hydrodynamic sequence.
Section~\ref{sec:results} presents
the results of the density reconstructions and ability to extract RMI growth rates using the ML architectures. 
%Section~\ref{sec:RMIUncertainty} presents an examination of the impact of uncertainty in capturing the radiographic features.  
Finally section~\ref{sec:Conclusions}
presents discussion of the simulation results and conclusions drawn from the numerical experiments.

\section{Dynamic Radiography Problem Setup}
\label{sec:ProblemSetup}

\newcommand{\xv}{\mathbf{x}}

Our objective is to recover the intricate details of the gas-metal interface by first estimating 
a time-series of densities $\left \{ \density_{t, i}(x) \right \}_{t=t_1}^{t_N}$, with $x \in \mathbb{R}^{3}$ and $i$ representing a material index,
from their corrupted radiographs $\left \{ {m}_t \right \}_{t=t_1}^{t_N}$, where ${m}_t \in \mathbb{R}^{N_1\times N_2}$ is a monochrome image with $N_1 \times N_2$ pixels.
The times $t_1, t_2, ..., t_N$ denote $N$ time points at which the radiographic measurements are collected.
In this investigation we consider the axisymmetric problem, where density objects are fully characterized by the central slice through the near-spherical object.  Following the determination of the density fields we then extract the gas-metal interface as described in Section~\ref{sec:results}.
%(i.e., essentially 2D densities).

%the density field $\density$. 
 %are made in the Abel domain. 
Our imaging model may be described as follows: %(see~\cite{mccann_local_2021} for further details).
The areal density of the object along a ray $r$ connecting the source and detector is denoted
%\add{\begin{equation} \label{eq:areal}
%    \rho_{A_i}(r) = \int_{-\infty}^\infty \rho_i(r_x(u), r_y(u), r_z(u)) du,
%\end{equation}}
\begin{equation}
\rho_{A_i}(r) = \abel[\density_i] = \int_{-\infty}^\infty \density_i(r_x(u), r_y(u), r_z(u))\, \mathrm{d}u,
\label{eq:abel}
\end{equation}
where $\abel$ is the forward operator corresponding to the Abel transform, $(r_x(u), r_y(u), r_z(u))$ is a parameterization of the ray $r$, $\rho_i(\cdot)$ denotes a time snapshot of a spatially varying density, and
$i$ is an index corresponding to the material being imaged.
Using a simple measurement model of a mono-energetic X-ray source,
%and a single material in the object is that
the number density of photons reaching the detector along ray $r$ is approximately given by
%\begin{equation}
%\label{eq:beer}
%    I(r) = I_{0}
%   \exp \left( - \xi \rho_{A}(r)\right),
%\end{equation}
\begin{equation}
\label{eq:beer}
    I(r) = I_{0}
    \exp \left( - \sum_{i} \xi_i \rho_{A_i}(r)\right),
\end{equation}
where  $I_0$ is the number density of the incident beam and $\xi_i$ is the mass attenuation coefficient of material $i$~\cite{berger_xcom_2010}.
A discrete, finite radiograph is measured in practice representing a finite grid of detectors.
This direct radiograph (without scatter) $\direct$
at each detector pixel is approximated as
\begin{equation} \label{eq:direct-mono}
    \direct[m, n] 
    = \int_{R_{m,n}} I(r)\, \mathrm{d}r
    \approx C I(r_{m,n})
\end{equation}
where $R_{m,n}$ denotes the rays impinging pixel $(m,n)$,
$r_{m,n}$ is the ray through the pixel center,
and
$C$ is a constant that depends on factors such as the detector pixel area.

The transmission (noisy radiograph)
includes contamination from several noise terms, and is given by
\begin{align}
    T_t = d_t + n_t,
\end{align}
where $d_t$ and $n_t$ are the direct 
radiograph and noise at time $t$.
At each time, we model the noise as
\begin{align}
    \noise &= D_{\rm dsb} + D_{\rm s} + B_{\rm s} + \eta
\end{align}
where $D_{\rm dsb}$ represents the blur from the 
source and detector, given by 
\begin{equation}
D_{\rm dsb} = D_{\rm sb} \ast  \phi_{\rm db}, \qquad
 D_{\rm sb} = \direct \ast G_{\rm blur} (\sigma_{\rm blur}).
\end{equation}
Here, $\phi_{\rm db}$ is a custom kernel associated with the 
detector and
$G_{\rm blur}(\sigma_{\rm blur})$ is a 2D Gaussian kernel
with standard deviation $\sigma_{\rm blur}$.
In addition, the noise model includes correlated 
scattered radiation, $D_{\rm s}$, and an uncorrelated linear tilted background 
scatter field, $B_{\rm s}$, given by
\begin{equation}
 D_{\rm s} = \kappa  \direct \ast G_{\rm scatter} ({\sigma_{\rm scatter}}), \qquad
 B_{\rm s} = 
 % \sum_{i=0}^{n} 
 a {x} + b {y},
\end{equation}
where $G_{\rm scatter}(\sigma_{\rm blur})$ is a 2D Gaussian kernel
with standard deviation $\sigma_{\rm scatter}$,
$a$ and $b$ are scalar constants, and 
${x}$ and ${y}$ are projections of the 
2D functions $x$ and $y$ to the image plane.
Finally, both gamma and photon noise are added as
\begin{equation}
\eta = \eta^{\rm Po}(\gamma_{\rm g}) \ast \phi_{\rm g} 
+ \eta^{\rm Po}(\gamma_{\rm p}) \ast \phi_{\rm p}.
\end{equation}
where $\eta^{\rm Po}$ is Poisson Noise and
$\gamma_{\rm g}$ and $\gamma_{\rm p}$ are the rates
and $\phi_{\rm g}$ and $\phi_{\rm p}$ are the 
gamma and photon kernels, respectively.
%The final observation is referred to as the total transmission
%and the actual observation is a corresponding total signal %$\transmission$ 
%$\m$
%a \emph{scatter}  signal $\scatter$ and additional noise $\noise$.
%as there can be contamination resulted from \emph{scatter} signal $\scatter$. 
%(i.e., $\m = \direct + \scatter + \noise$).
%In the CT setting, the measurement is made in the form of radiography. 

%The scatter signal is modeled using 
%$\scatter_t = \beta \mathcal{G}[\direct_t]$
%where $\mathcal{G}[\cdot]$ denotes convolution with a Gaussian filter and $\beta$ is a scaling capturing the strength of the scatter field relative to the direct signal.
The choice to model scatter as a kernel convolved with the direct signal
is common in the scatter correction literature~\cite{sun_improved_2010,mccann_local_2021}.
This approach provides a fast scatter model that is representative of models used in practice.
%in practice.
%The additional noise, $\noise_t$ is assumed to be Gaussian noise, consistent with 
%$\noise_t$ is typically assumed to be a stochastic field (e.g., Poisson or Gaussian noise)~\cite{thibault2006recursive,lu2001noise,demirkaya2001reduction,sauer1991nonstationary}.
%$\scatter_t$ 
%is typically assumed to be a function
%of $\direct_t$ or $\density_{t, i}$.
%Commonly, $\scatter_t$ is modeled as $\scatter_t = \mathbf{k} \ast \direct_t$, obtained as the convolution of the direct radiograph with a scatter kernel $\mathbf{k}$, which can be a Gaussian kernel or a kernel that depends on the underlying density distribution~\cite{sun_improved_2010,mccann_local_2021}. 
%$\scatter_t$ 
%
The above model can be readily extended to polyenergetic X-ray sources~\cite{elbakri_statistical_2002,mccann_local_2021}. 
This work focuses on material objects that are of interest in double shell ICF capsules,
and we work with monoenergetic X-ray sources for simplicity. 
%\scatter = \gaussian(\direct), 
%$\gaussian$ denotes the underlying scatter function, and $m$ is a random scaling factor with the uniform distribution on $[0.95, 1.05]$.

Great difficulty in performing the density time-series reconstruction arises due to the presence of scatter, noise, and additional factors including 
the variability of the beam spot, energy spectra, and model mismatch in the forward operator.
Incorporating a time-series for the evolution of the underlying density fields over time
could provide improved dynamic reconstructions~\cite{myers2011dynamic,bonnet2003dynamic,desbat2007algebraic,jailin2018dynamic,hossain2022high}. 
To this end we adopt a machine learning approach to incorporate hydrodynamic priors to encourage the estimated density time-series to be consistent with the equations of motion that govern their evolution.

The evolution of the density over time in a dynamic experiment can be modeled by a system of partial differential equations (PDEs) describing radiation hydrodynamics.  To facilitate the analysis of the radiation-hydrodynamic system we utilize the Euler equations in a manner analogous to Bello-Maldonado. \cite{toro2013riemann, bello2020matrix} 
%Utilizing these equations in the reconstruction process would   %\add{By utilizing these equations, the correlations between the temporal projections may be imposed on the reconstruction so that the collectively reconstructed dynamics is consistent with the equations of motion that govern the evolution of the inferred field.}
 The system of PDEs comprise a continuum model governing density evolution and is applicable in many realistic scenarios where dynamic radiography is applied. 
%The governing equations are given by
%\begin{gather}
%    \partial_t\rho + \nabla\cdot(\mathbf{u}\,\rho) = 0,\quad \rho\,(\partial_t\mathbf{u} + (\mathbf{u}\cdot\nabla)\mathbf{u}) = -\nabla p,\quad \rho\,(\partial_t e  +\mathbf{u}\cdot\nabla e)=-p\,\nabla\cdot \mathbf{u},
%\end{gather}
%where $\rho$ is the mass density, $\mathbf{u}$ is the fluid velocity, $e$ is the specific internal energy, and $p = P(\rho,e)$ is the pressure expressed as a point-wise function of $\rho$ and $e$. 
A closure of the hydrodynamic variables in the form of a material-dependent equation of state (EOS)
% The precise form of $P(\rho,e)$ is determined by a fluid's equation of state (EOS), 
is required, along with appropriate initial and boundary conditions, to uniquely prescribe the time evolution of any hydrodynamic variable, including the density.
In practice, the EOS and other conditions or parameters for a specific experiment or test case are usually unknown.  
In this case, directly using the PDEs to enable dynamic reconstruction is not feasible.
In our study, we numerically integrate the Euler equations with 
varying choices of EOS model parameters
to generate a large dataset of density time-series
and learn an EOS-agnostic model for reconstructing densities.

%priors to enable accurate density estimation as discussed in Section~\ref{sectionmethod}.

%Each set of clean data of $\density$ contains 41 frames, which corresponds to 41 time points separated by time intervals of the same length, and describes an instance of density evolution in that time window. One main challenge in our study is to capture the dynamics behind the time series data. This motivates us to add an unsupervised component in our training scheme.

%\paragraph{Remark} The noise/scatter existing in the input density time-series is introduced in the radiograph domain. The first noise source is the $\scatter$ term in the transmission signal and the second noise source may be white noise. 

\section{Hydrodynamic Test Problem}
\label{sec:DataGeneration}

\subsection{Generation of Density Time Series}
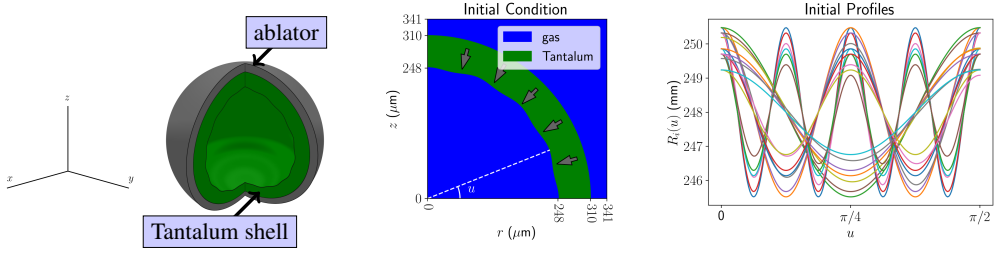
\begin{figure}%[htbp]
  \centering
  \resizebox{\columnwidth}{!}{%
  \begin{tikzpicture}
  \useasboundingbox (0,0.25) rectangle (15,4);  
  
  \draw(.5,0) node[anchor=south] {\trimh{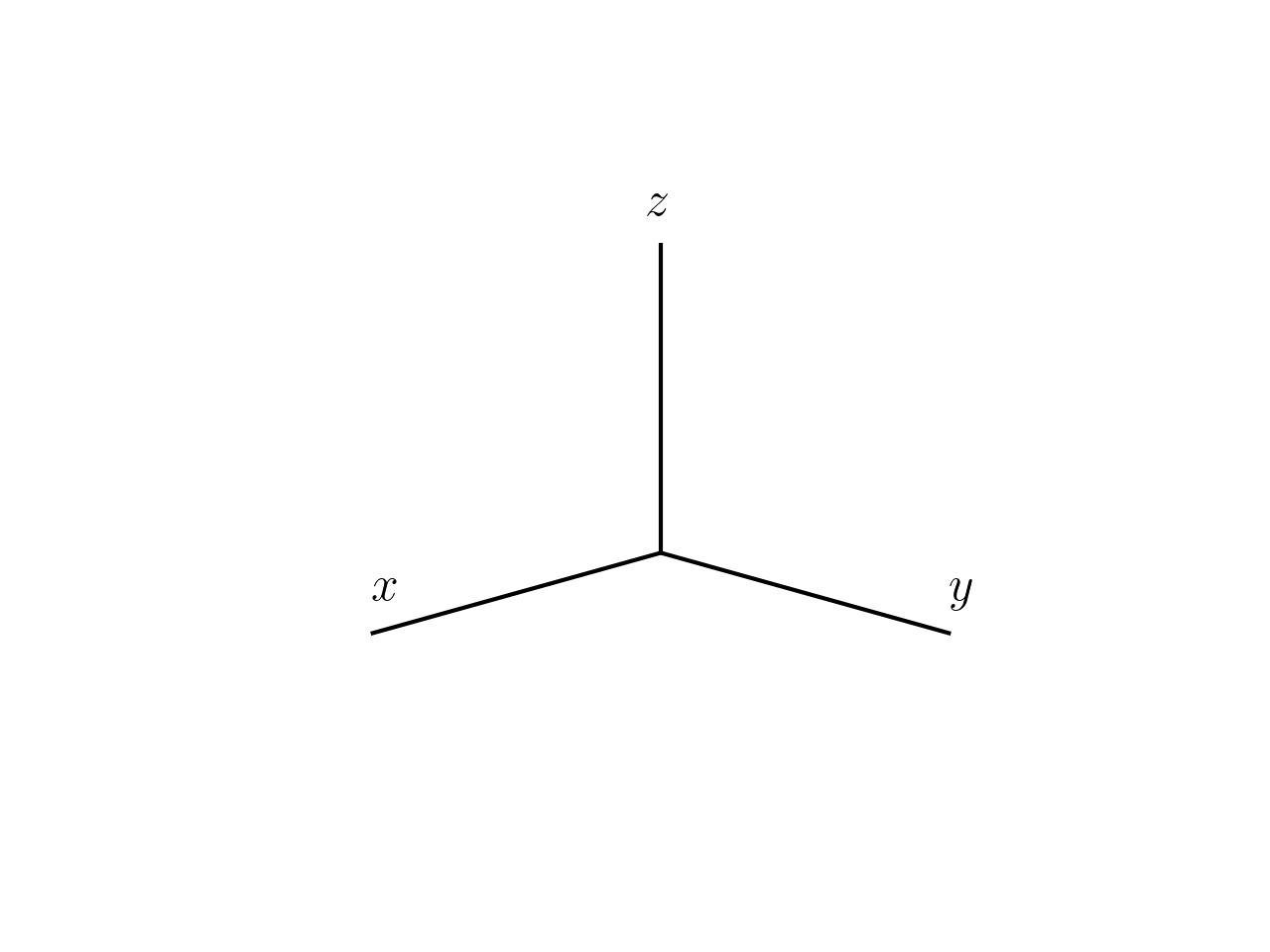}{3 cm}{0}{0}{0}{0}};
   \draw(3.25,0) node[anchor=south] {\trimh{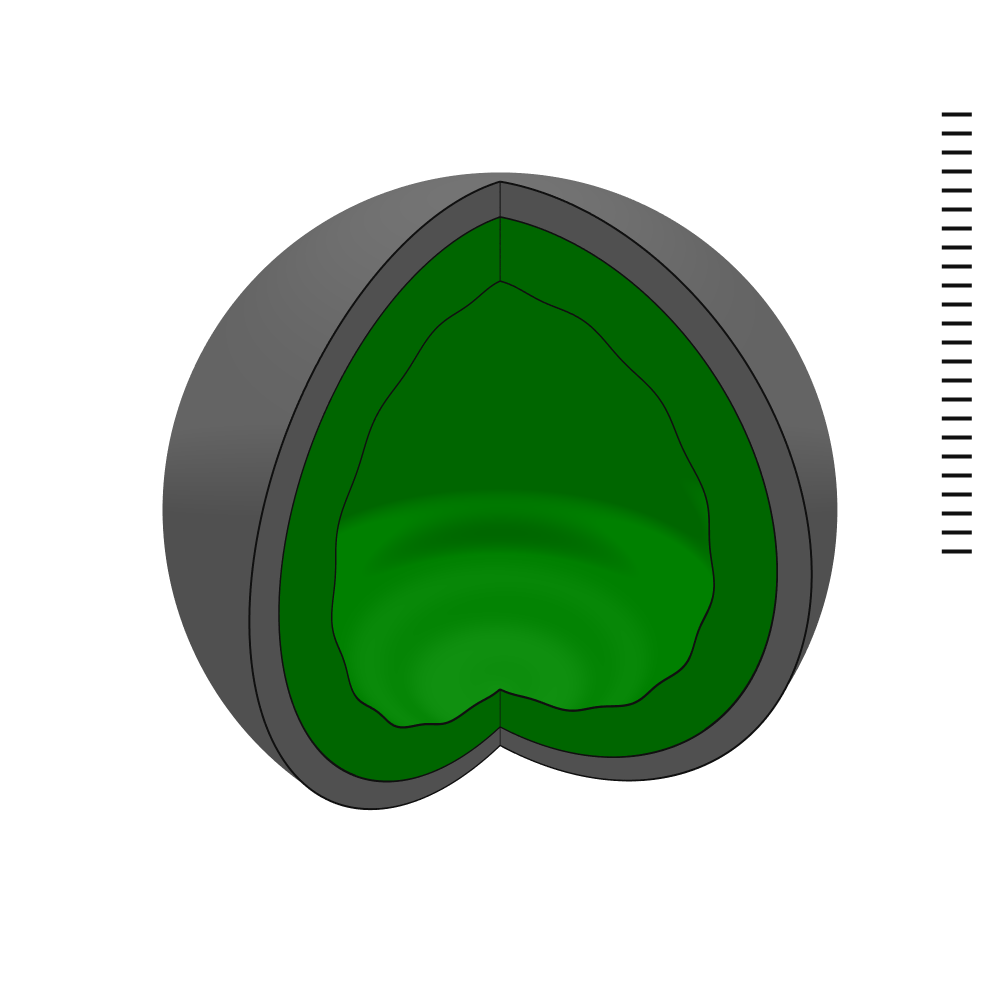}{3.5 cm}{.1}{.1}{0}{0}};
   \draw(10.25,0) node[anchor=south east] {\trimh{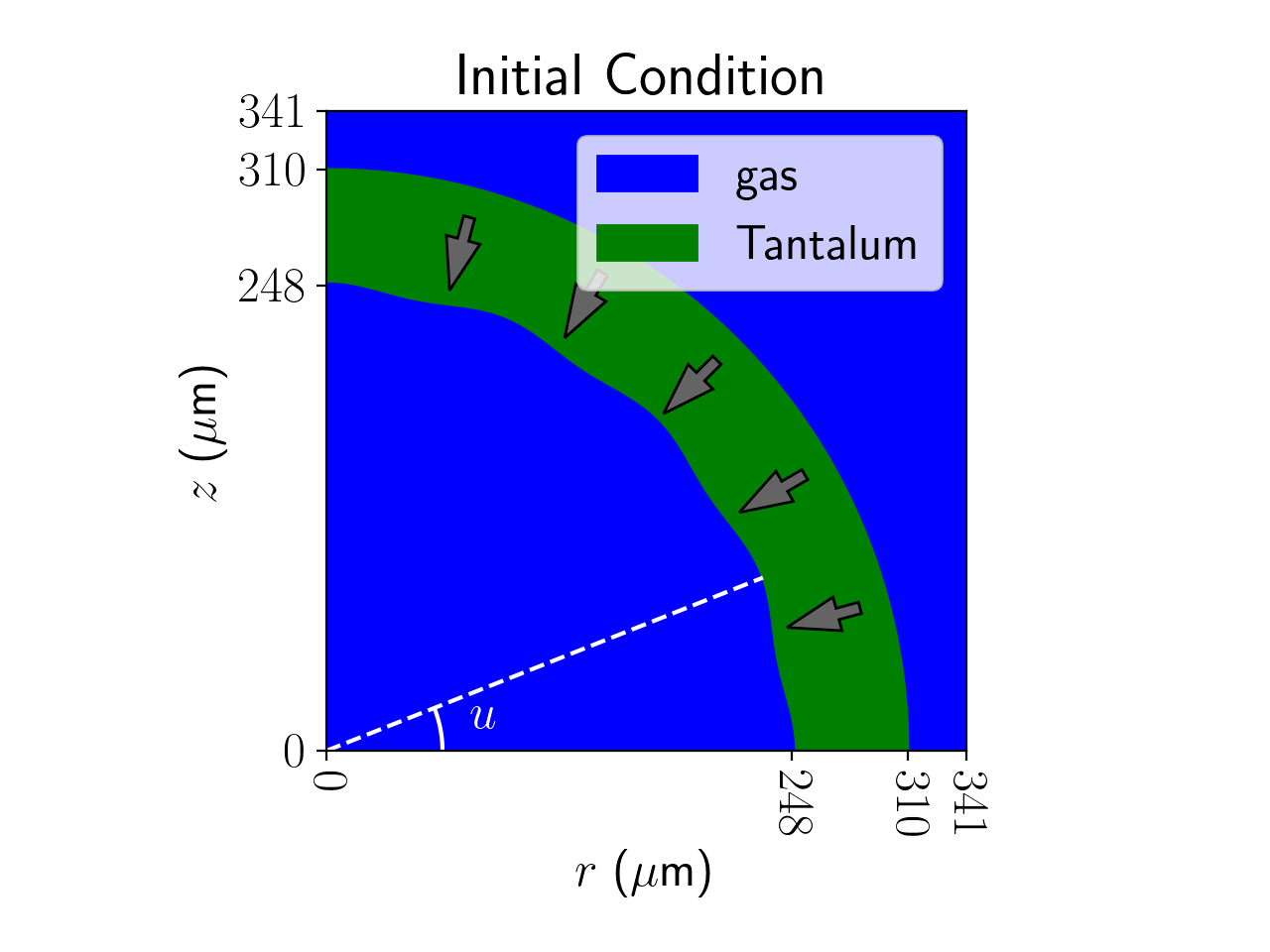}{4 cm}{.05}{.0}{0}{0}};
   \draw(15,0) node[anchor=south east] {\trimh{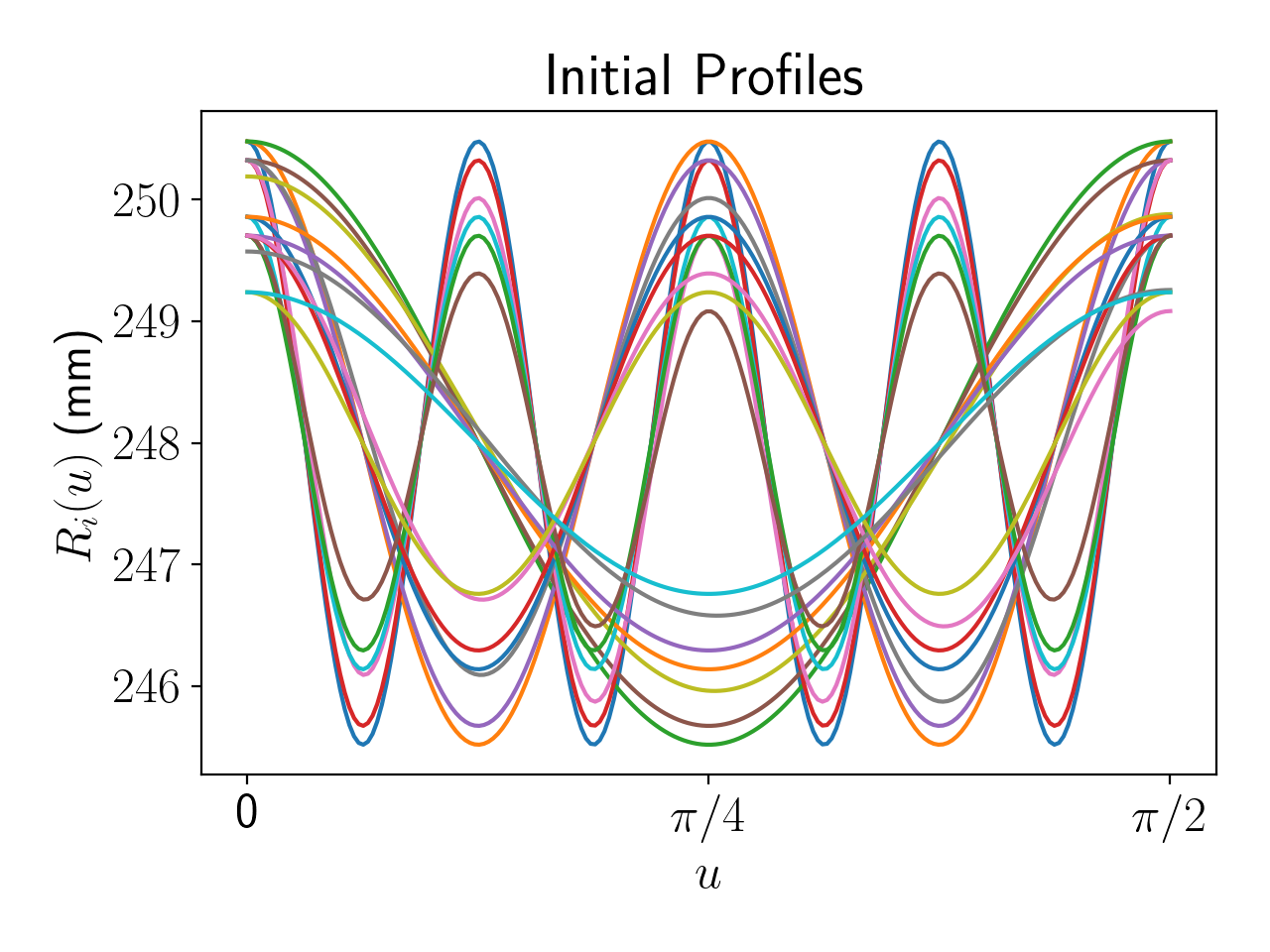}{4 cm}{.0}{.0}{0}{0}};

   \coordinate (A) at (4, 3.5);
   \coordinate (B) at (3, .5);
   \draw[->, line width=1.5 pt] (A) -- (3.5, 3);
   \draw[->, line width=1.5 pt] (B) -- (3.6, 1.1);
   \draw (A) node[draw, fill=blue!20] {ablator};
   \draw (B) node[draw, fill=blue!20] {Tantalum shell};

  \end{tikzpicture}  
  }
  \caption{Left: 3D mock-up of a Tantalum shell (green) with a
  perturbation on the interior surface and an outer ablator layer (gray).
  Middle: projection of the Tantalum shell onto $(r, z)$ 
  coordinates. The inner radius is parameterized by
  the angle, $u$, between the white dotted line and
  the $r$ axis. 
  The drive from the ablator is modelled as an initial velocity 
  on the Tantalum shell ($v_{\rm impl}$).
  Right: Plot of the 20 separate profiles for 
  radius of the perturbed inner surface verses angle $u$. 
  }
  \label{fig:initialcondition}
  %\vspace{-0.1in}
\end{figure}

As a test problem, we study shock propagation in a time-dependent (3D) density profile,
created by an implosion of a nearly-spherical ICF double shell configuration.
The study is limited to 
azimuthal symmetry so that the density at any time can be described in cylindrical coordinates~$(r, z)$. 
Additionally we restrict our attention to the Mie-Gr\"uneisen (MG) EOS model.
Simulations are performed 
on a $440\times 440$
uniform Cartesian grid on a 
computational domain given by the quarter-plane $[0, L] \times [0, L]$, where 
$L=341$~$\mu$m.
The uniform grid cell size is $\Delta r = \Delta z = \frac{440}{L}$.
The metallic shell is made of Tantalum and its density is initially uniform at a value of 16.65~g/cc.
The inner surface of the Tantalum can be described as the set of 
coordinates $(r_{\rm in}(u), z_{\rm in}(u))$ satisfying 
\begin{align}
(r_{\rm in}(u)^2 + z_{\rm in}(u)^2)^{1/2} = R_{\rm in} 
+ \sum_{k=1}^{8} F_k \cos(2k u), \qquad
u \in [0, \pi/2],
\end{align}
where ${R_{\rm in} = 248}$ $\mu$m, $F_k$, $k=1,\dots,8$, are coefficients of the perturbation corresponding to the $k^{\rm th}$ cosine harmonic.
The outer surface of the shell is a sphere with radius ${R_{\rm out} = 310}$ $\mu$m.
There are 20 different inner surface perturbation profiles considered in our dataset. The
corresponding coefficients are recorded in Table~\ref{tab:initialcoeffs}.
Figure~\ref{fig:initialcondition} presents an initial perturbation given to the interior shell.
As an initial condition, the shell is given a uniform implosion velocity, $v_{\rm impl}$,
in the direction of the origin to initiate an implosion.
%in our dataset as indicated in Table~\ref{tab:vary}.

% % /-------------------------------------------------------------------------------\

% % \-------------------------------------------------------------------------------/

In addition to varying the profile of the inner radius and the initial implosion velocity, our dataset consists of simulations covering 
parameters characterizing the MG equation of state~\cite{hertel98b},
\begin{align}
    p\left(\chi, T\right) 
    = \frac{\rho_0 c_s^2\chi\left(1 - \frac12\Gamma_0\chi\right)}
          {(1 - s_1\chi)^2}
    + \Gamma_0\rho_0 c_V (T - T_0),
\end{align}
where $\chi=1-\frac{\rho_0}{\rho}$,
$\rho_0$ and $T_0$ are the reference density and temperature, respectively, $c_s$ is the speed of sound, $\Gamma_0$ is the Gr\"uneisen parameter at the reference state, $s_1$ is the slope of the linear shock Hugoniot curve, and $c_V$ is the specific heat capacity at constant volume.
Out of these parameters, we keep the reference density $\rho_0$ fixed at 16.65~g/cc and the reference temperature $T_0$ fixed at 0.0253~eV. The parameter set $\{c_s,\, s_1,\, \Gamma_0,\, c_V\}$ is varied as shown in Table~\ref{tab:vary}. 
%The parameter values used in this examination are provided in  Table~\ref{tab:nominal}.
%\begin{table}[htbp]
 % \centering
%  \begin{tabular}{lcccccc} 
%  \hline \hline
 % Material & $\rho_0$ [gm/cc] & $T_0$ [K] & $c_0$ [km/s] % & $s_1$ & $\Gamma_0$ 
%            & $c_V$ $[{\rm erg}\;{\rm g}^{-1}\;{\rm % eV}^{-1}]$\\
%  \hline
 % Tantalum    & 16.65 & 293.15 & 3.39 & 1.22 & 1.6 & %$1.6\times10^{10}$ \\
%  \hline \hline
%  \end{tabular}
%  \caption{Parameter values in the employed Mie-Gruneisen %(MG) equation of state. The reference density parameter %$\rho_0$ is fixed.}
%  \label{tab:nominal}
% \end{table}

\remove{
Finally, in hydrodynamic computer codes %such as CTH, 
the stress-tensor components are split into a hydrostatic equation of state and a modified elastic-perfectly plastic constitutive model. In these simulations, we utilize a Preston, Wallace, Tonk (PTW) strength model applicable to metals at high strain rates.  Table~\ref{tab:vary} presents the available choices for shear modulus in our dataset.
}

\begin{table}[htbp]
  \centering
  \begin{tabular}{|c|ccccc|} 
  \hline 
    Options  & 1   & 2  & 3 & 4 & 5 \\
  \hline
     %$v_{\rm impl}$ [m/s]  & 950   & 943.35  & %946.20 & 959.50 & 954.75 &\\
  % \hline
    $\Gamma_0$  & 1.6   & 1.7  & 1.76 & 1.568 & 1.472 \\
  \hline
    $s_1$  & %1.32   
    1.22 & 1.464  & 1.342 &  & \\
  \hline
    $c_s$ [m/s]  & 339000   & 372900  & 305100 & 355000 &  \\
  \hline
    $c_V$ $[{\rm erg}\;{\rm g}^{-1}\;{\rm eV}^{-1}]$  & $1.6\times 10^{10}$   & $1.76 \times 10^{10}$  & $1.44 \times 10^{10}$ &    &  \\
  \hline
  \end{tabular}
  \caption{Matrix of parameter values used to develop the simulated dataset. All combinations of above parameters are used to simulate our data.
  }
  \label{tab:vary}
\end{table}

\begin{figure}%[H]
    \begin{subfigure}{\linewidth}
        \centering
        \includegraphics[trim=3cm 1cm 3cm 0, width=.65\textwidth]{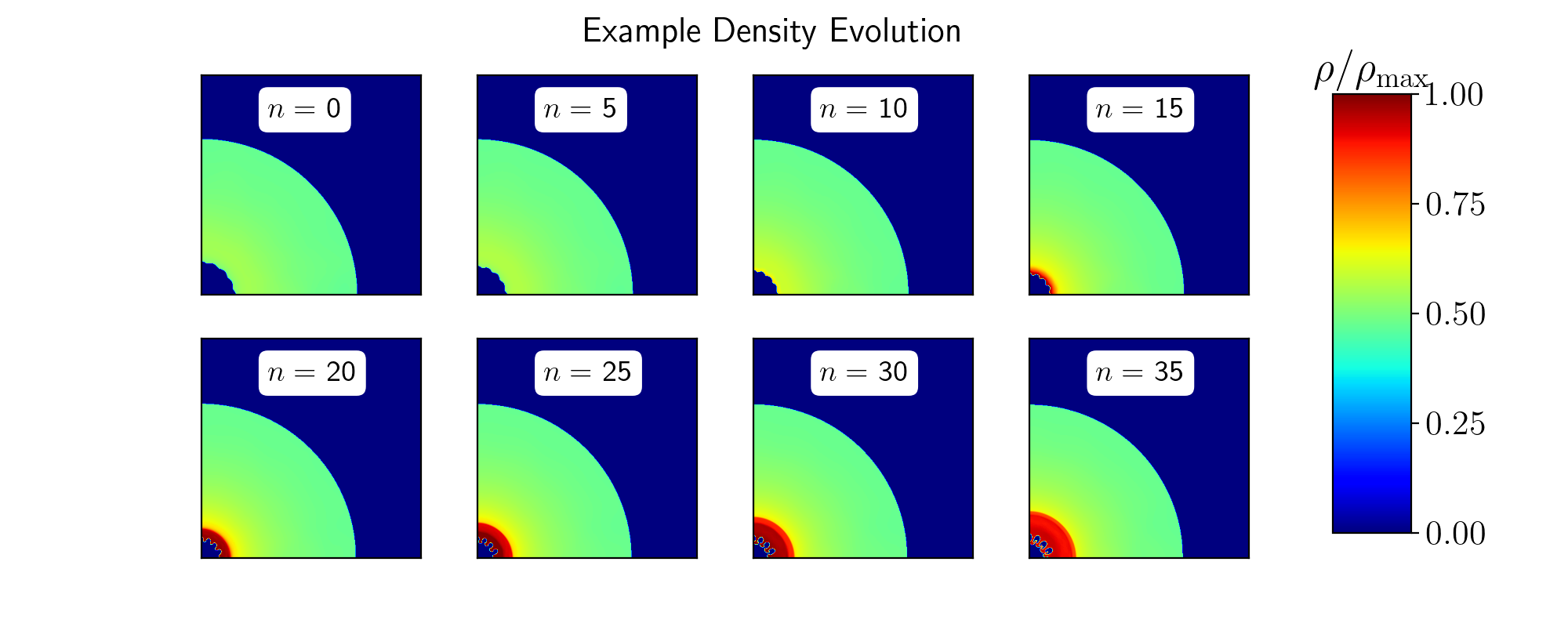}
        \subcaption{Example of the density evolution for an initial condition with inner surface perturbation profile 1 for time indices $0, 5, \dots, 35.$
        % in between 
        % $t=.37$ $\mu$s and $t=.405$ $\mu$s. 
        The images are 440x440 pixels representing the domain $[0, L] \times [0, L]$.}
        \label{fig:dyn_gt_type1}
    \end{subfigure}
    \vfill
    \begin{subfigure}{\linewidth}
        \centering
        \includegraphics[trim=3cm 1cm 3cm 0, width=.65\textwidth]{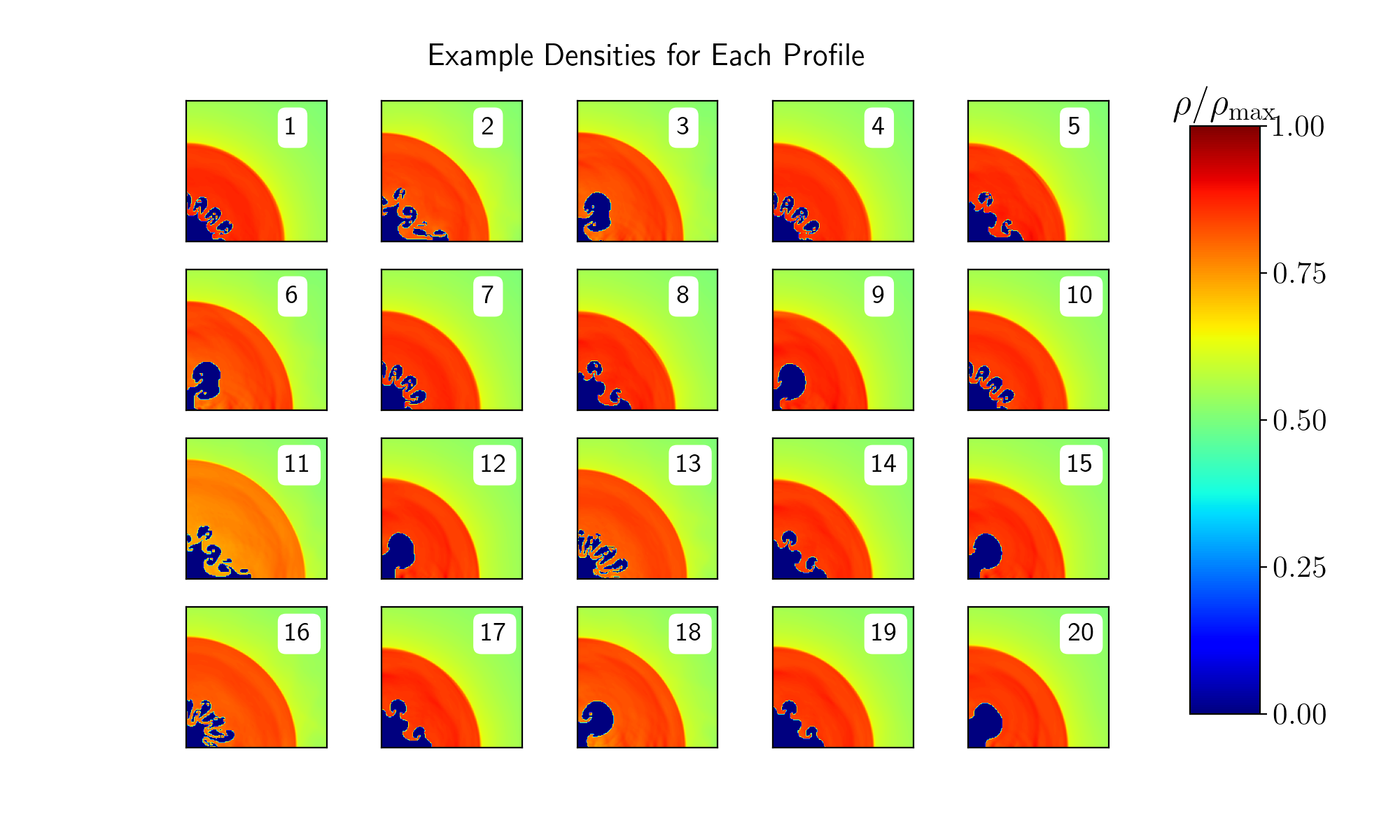}
         \subcaption{Examples of the density at time index 40 for each inner surface perturbation profile. The images are 150x150 pixels representing the domain $\left[0, \frac{15}{44} L\right] \times \left[0, \frac{15}{44} L\right].$}
        \label{fig:dyn_gt_type2}
    \end{subfigure}  
    \caption{Example plots of the density evolution (a) and the various RMI profiles representing each inner surface perturbation profile (b).}
    \label{fig:dyn_egs}
\end{figure}

Altogether, the dataset realizes every unique parameter combination in a 7-dimensional parameter cube with $28,800$ total simulations. 
Each hydrodynamic simulation is comprised of 
density field snapshots 
% at times $(0.37 + 0.001 n) \,\mu{\rm s}$, $n=0,1,\dots,40$.
at later times when the instability is present. We label these times as $n=0, 1, \dots, 40$.
An example of a density time series is shown in
Figure~\ref{fig:dyn_egs}.
Once the Tantalum shell has collapsed, a shock is formed and reflected from the axis.  The shock then interacts with the perturbed inner Tantalum edge. This creates an RMI.  The topology of this interior evolves as depicted in Figure~\ref{fig:dyn_egs}. The expanding shock proceeds to propagate into the
non-constant dynamic density background. 
We chose frames corresponding to the time instants at $n=25, 30, 35, 40$
to train the network in our studies. 

\subsection{Generation of Synthetic Radiographs}

Synthetic radiographs are produced at each time step.
The \texttt{create\_sino\_3d} function from the
ASTRA Toolbox~\cite{astra} is used to evaluate the forward operator 
in~\eqref{eq:abel}.
Equation~\eqref{eq:beer} is used to obtain the direct radiograph signal
from the areal mass using
$I_0=3.201 \cdot 10^{-4}$,
$\xi_{(\rm gas)}=9.40\;\text{cm}^2/\text{g}$, $\xi_{(\rm Tantalum)}=\xi_{(\rm coll)}=13.03\;\text{cm}^2/\text{g}$.
The source blur kernel is a 2D Gaussian kernel
with $\sigma_{\rm blur}$ chosen randomly 
between 1 and 3 pixels
with a random orientation between 5 and 26 degrees.
The correlated scatter kernel is also a 2D Gaussian kernel 
with $\sigma_{\rm scatter}$ chosen randomly between
10 and 30 pixels and scatter level, $\kappa$,
chosen randomly between 10 and 30.
The coefficients of the background scatter field are 
chosen such that the level is randomly between 0.5 and 1.5
times the mean signal level in the center of the image
and the tilt is between -10\% and 10\%.
The level of the gamma noise is randomly set in the range
(39,000, 50,000)
and the level of the photon noise is randomly set 
in the range (350, 450). 
Each random number is generated independently for each 
simulation and time step.

\begin{figure}[tb]
  \centering
  \includegraphics[trim=0 0 0 0, clip,
  width=.4\textwidth]{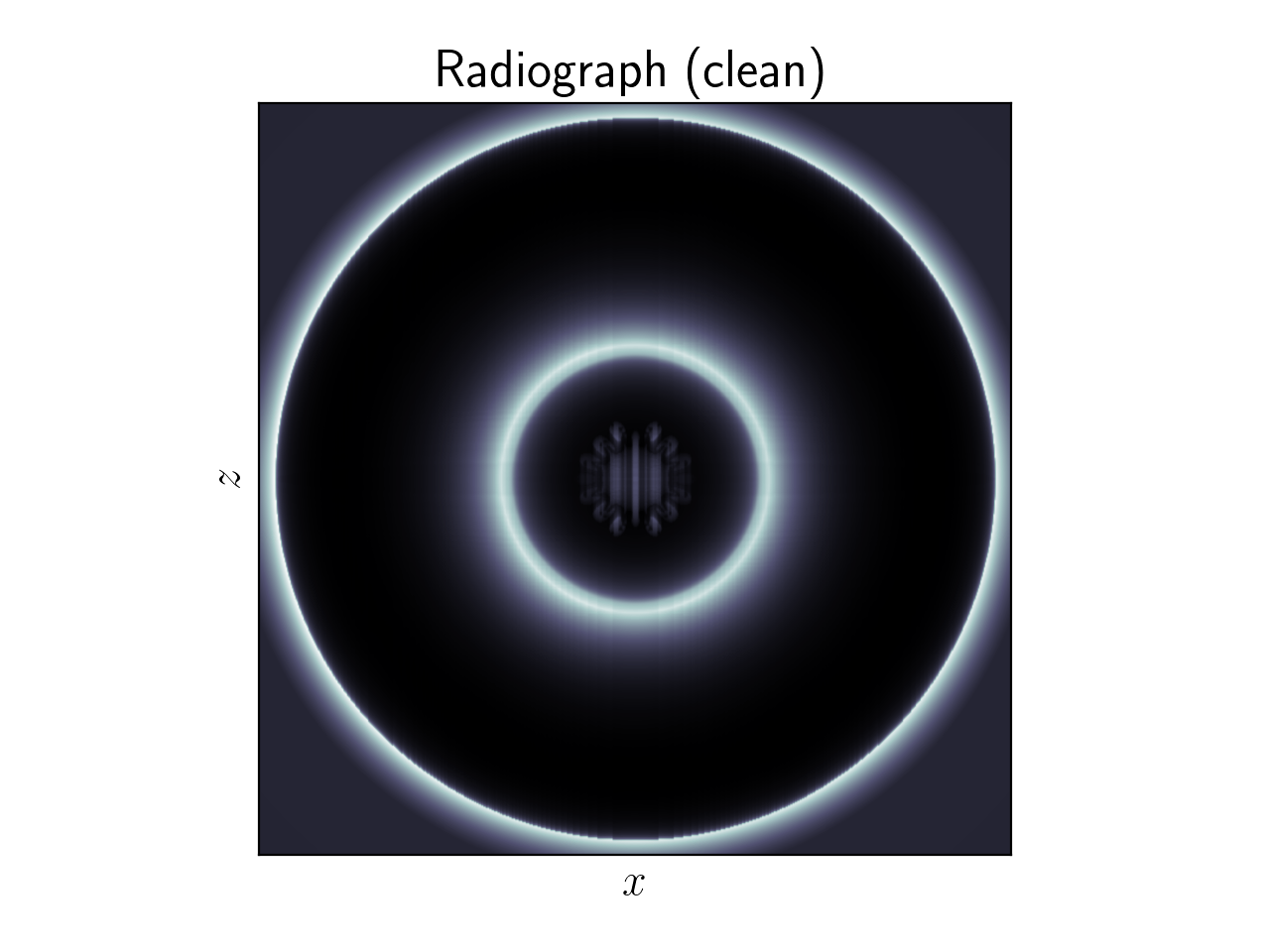}  
  \includegraphics[trim=0 0 0 0, clip,
  width=.4\textwidth]{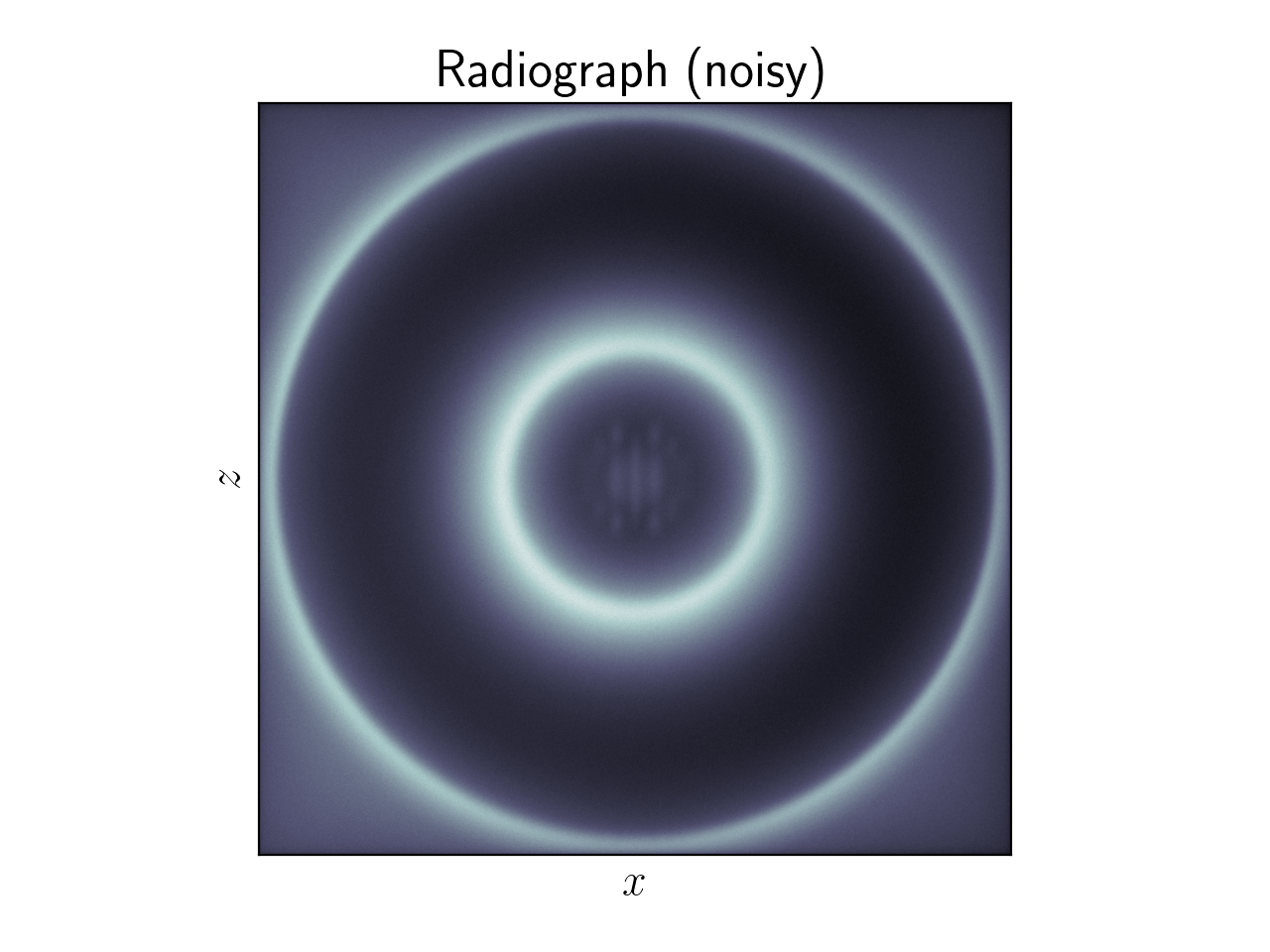}
  \includegraphics[trim=0 0 0 0, clip,
  width=.4\textwidth]{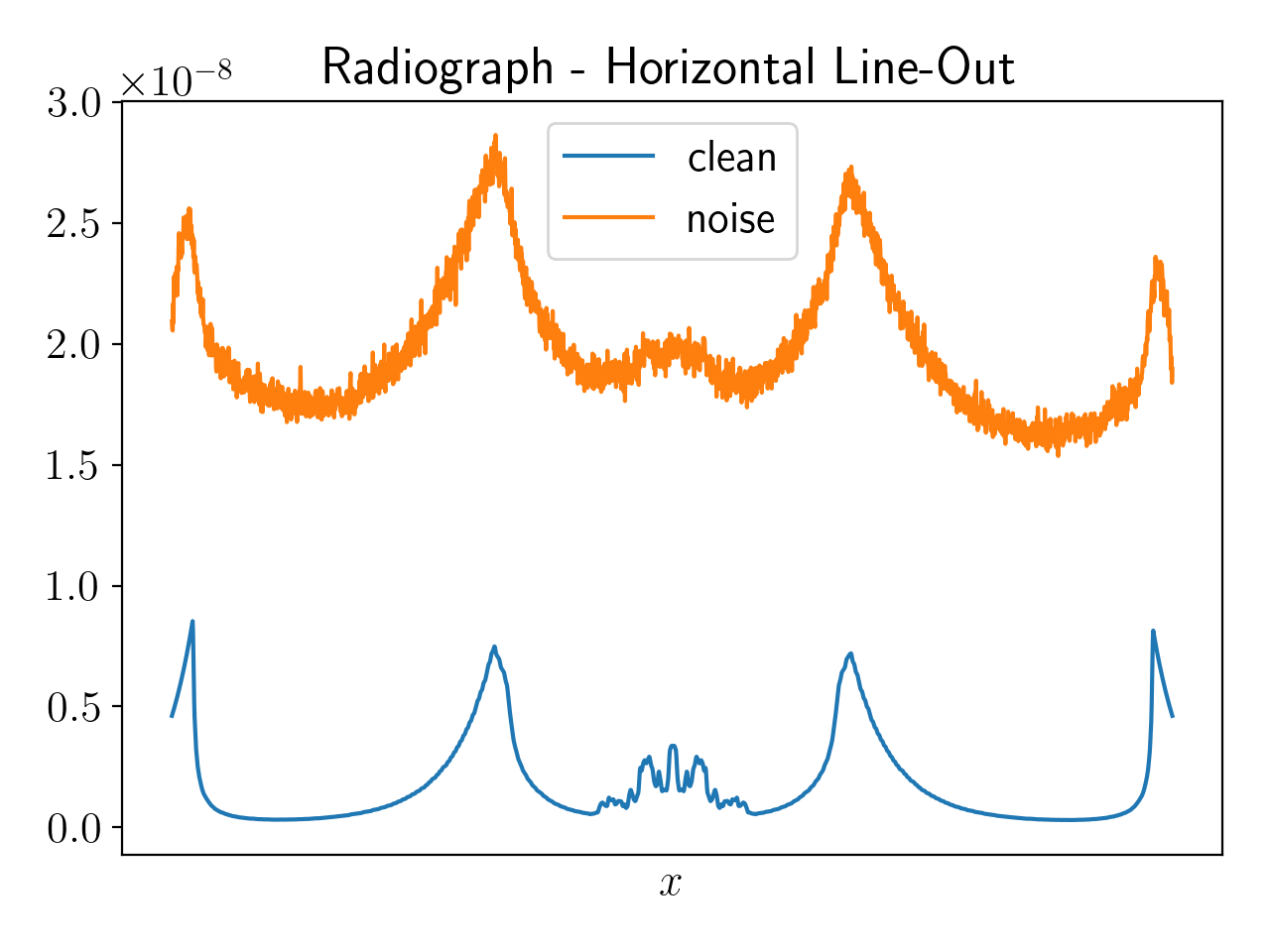}  
  \includegraphics[trim=0 0 0 0, clip,
  width=.4\textwidth]{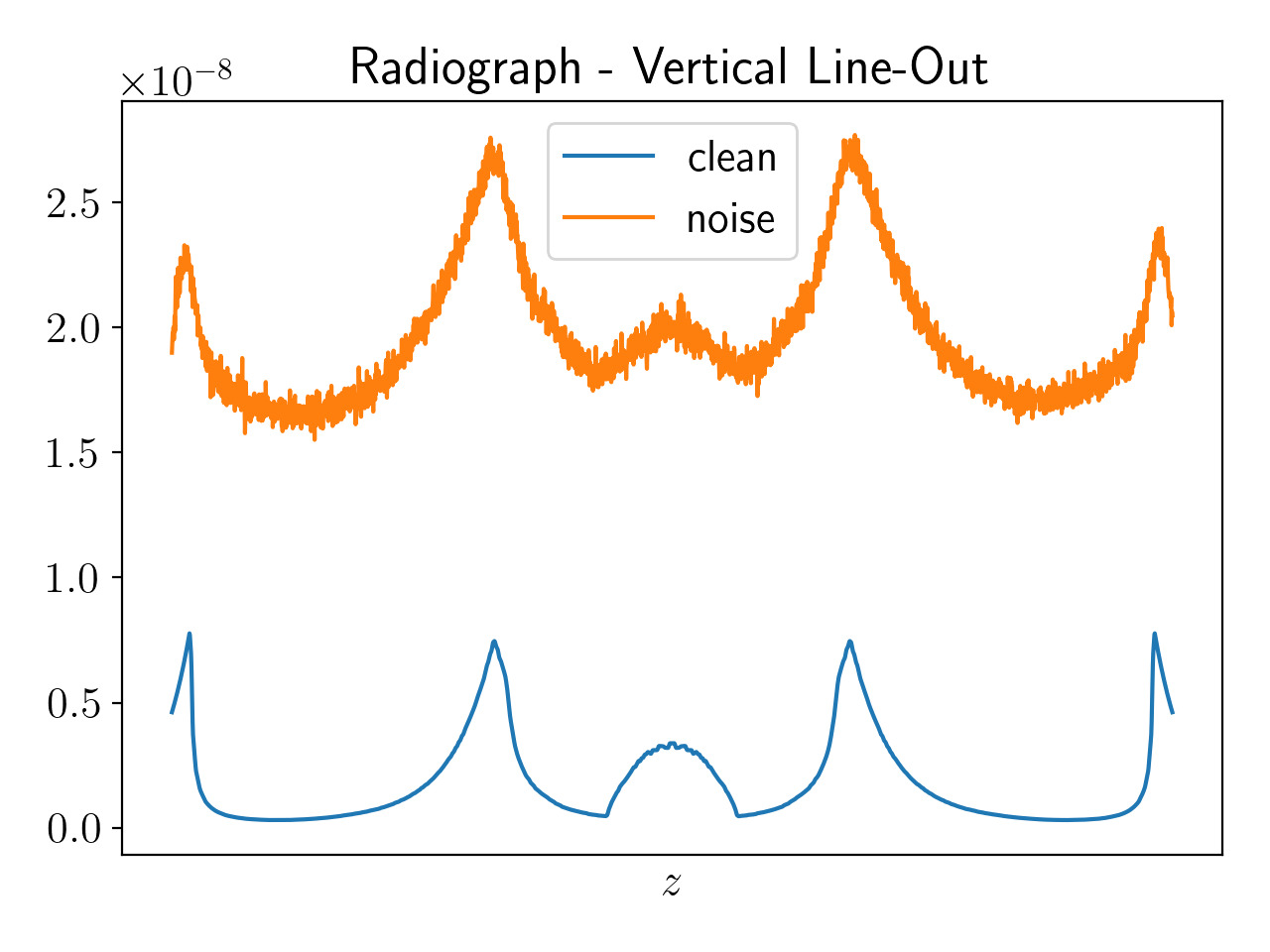} 
  \caption{Upper left: radiograph before applying noise. Upper right: radiograph after applying noise. Bottom left: horizontal line out through the center of the radiographs. Bottom right: vertical line out through the center of the radiographs. In both line plots, the orange line 
  corresponds the noisy radiograph and the blue line corresponds to the radiograph without noise.
  %\\
%\texttt{data\_ta\_2d\_profile1.vel0.mgrg00.s10.cs0.cv0.ptwg00.pkl}
  }
  \label{fig:sampleradiographsingleline}
\end{figure}

Figure~\ref{fig:sampledensitysingle} shows an example of a density field at
time index 40 and a synthetic radiograph, $\m$, generated using the above method.
For the same example, Figure~\ref{fig:sampleradiographsingleline} shows 
the direct signal, $\direct$, radiograph $\m$, 
and profiles across the center of the two images.

\subsection{Generation of Shock Features}

% Talk about the use of robust features here
One of the primary aspects of the ICF shell dynamics is the 
evolution of the inner gas-metal interface,
i.e., the growth of the instability. This is because the
passage of the incoming and outgoing shocks through this interface
renders it unstable to the RMI. Considering temporally evolving simulations, we
are interested in times when the instability on this interface has
permitted the growth of perturbations to the extent that the inner
gas-metal interface displays significant asymmetry. As such, we assume
that the interface as identified by the feature extraction procedure
is not robust. That is, we expect that the interface as identified by
the imaging and feature extraction procedures is sensitive to the dependent the measurement model. This is in contrast with the shock and
outer edge features that we assume are robust. The robustness of these
latter features is due to the symmetry of the setup and dynamics and
the stable nature of their evolution. Nevertheless, because of its
passage across the unstable inner gas-metal interface, we expect the
stably evolving shock to be imprinted with a decaying set of
perturbations that can be reliably identified. 

Accordingly, we have extracted shock and edge features
at each time for each sequence of density fields. 
An edge-detection algorithm was utilized so as to enable the determination of a parametric
representation of the shock and edge as a function of polar angle~\cite{hossain2022high}.
These features are subsequently compressed into a low-dimensional 
representation in terms of cosine harmonic coefficients,
\begin{align}
r^{(i)}(\theta) = \sum_{j=0}^{N^{(i)}} F^{(i)}_j \cos(2 j \theta) ,
\end{align}
for $i={\rm shock},{\rm edge}$. We found
that $N^{(\rm shock)}=8$ and $N^{(\rm edge)}=5$
can represent the shock and edge features with sufficient 
accuracy across the dataset.

\subsection{Noise Model for Feature Extraction}
\label{sec:noisemodel}
The focus of this paper is to develop a features-to-density network
to be used as the second component in a radiograph-to-features-to-density 
pipeline for the purpose of quantifying the growth rates of RMI in double shell ICF environments. 
For simplicity and modularity, we built each component independently,
where the feature extractor was trained on radiograph and feature data pairs
and the features-to-density network was trained on feature and density 
data pairs.
To test the performance of the trained features-to-density network,
we developed a noise model to simulate the errors 
that would arise from the feature extraction model.

The feature extraction model is 
a CNN
consisting of an image Fourier feature encoding (IFFE) layer, convolution layers, and fully connected layers.
The feature extractor is applied to a subset of the synthetic radiographs. 
Errors between the reconstructed cosine harmonic coefficients 
and that of the ground truth were calculated and used 
to compute a sample mean and covariance matrix for the shock.
%$\boldsymbol{\mu}^{(i)}\in \mathbb{R}^{N^{(i)}+1}$, and covariance matrix,
%$\boldsymbol{\Sigma}^{(i)}\in\mathbb{R}^{(N^{(i)}+1)\times (N^{(i)}+1)}$,
%for $i={\rm shock},{\rm edge}$.
Using these computed statistics, we simulated the error produced from reconstructing the cosine harmonic coefficients from 
synthetic radiographs using a multivariate Gaussian noise model.
During model testing,
we added noise consistent with this distribution
to the inputs to better characterize the performance of the model.

\newcommand{\bogus}[1]{}
\bogus{
In Figure~\ref{fig:mean_and_cov},
the left plot shows the distribution of each harmonic
and its corresponding mean reconstruction error;
the right plot shows the covariance matrix.

\begin{figure}[tb]
  \centering
  \includegraphics[trim=0 0 0 0, clip,
  width=.4\textwidth]{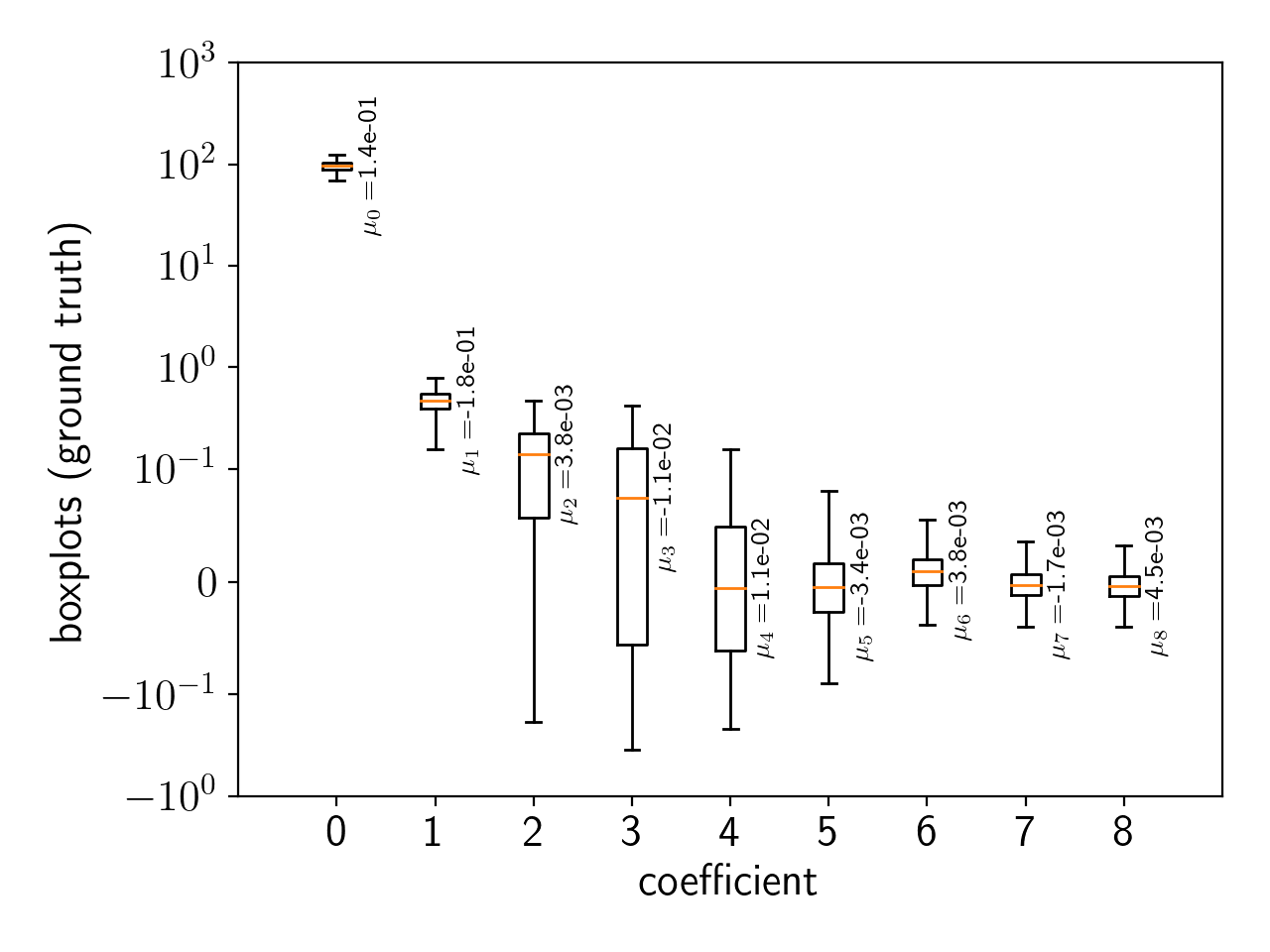}
  \includegraphics[trim=0 0 0 0, clip,
  width=.4\textwidth]{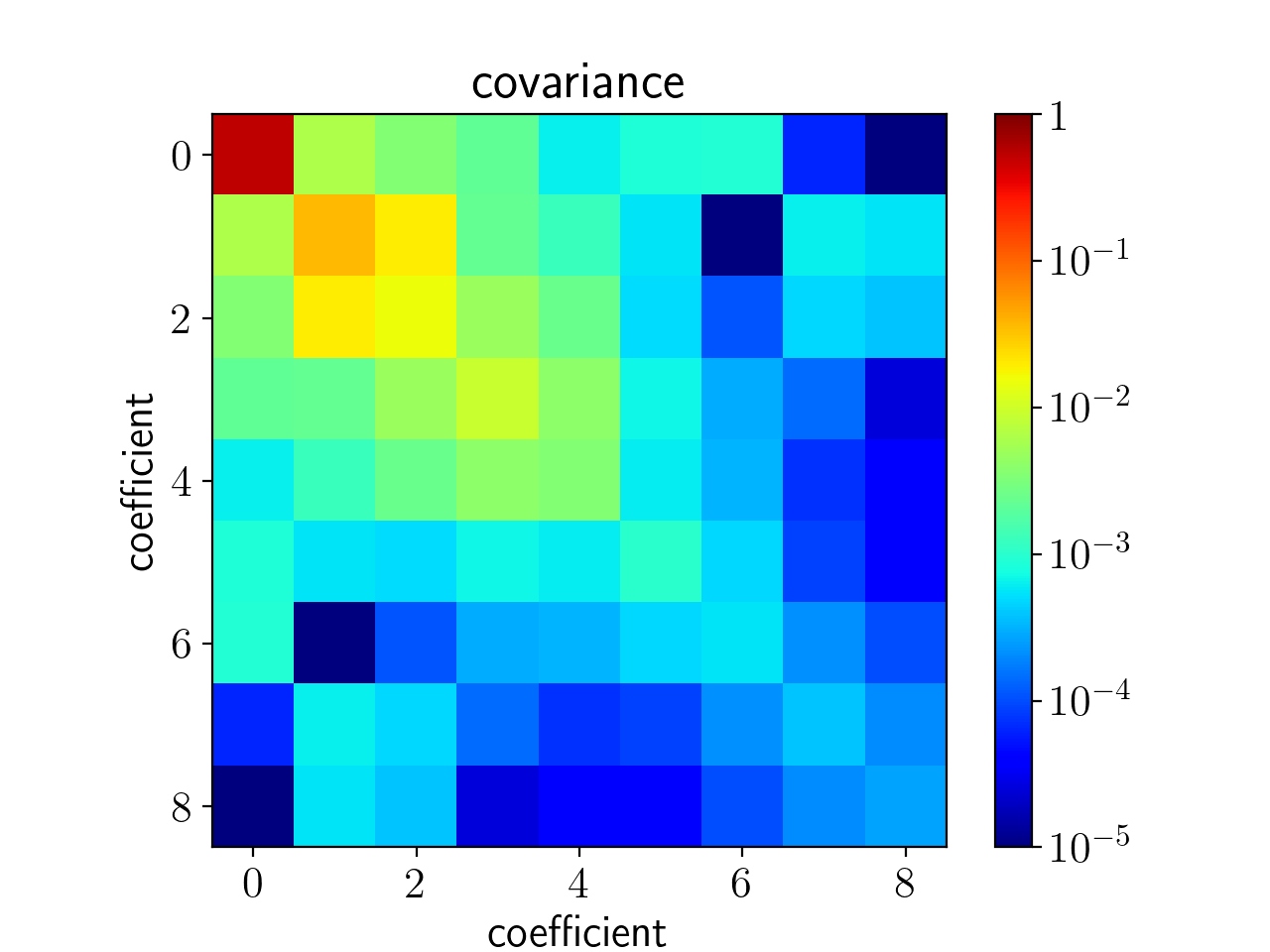}
  \caption{Left: box-plots depicting the 
  distribution statistics for each shock coefficient,
  labeled by the corresponding 
  mean radiograph-to-feature error.
  Right: covariance matrix of shock cosine coefficient errors on a log scale.
  }
  \label{fig:mean_and_cov}
\end{figure}  
}

%%%%%%%%%%%%%%%%%%%%%%%
\section{Description of Features-to-Density Architectures}
\label{sec:ArchitectureDescription}

We compared two different architectures 
for solving the features-to-density problem.
These approaches involve a generative
variational autoencoder (VAE) network based on
the vision transformer~\cite{dosovitskiy2021an},
which we refer to as the ShockDecoderViT, 
and a deterministic structure-preserving network
based on the original transformer~\cite{Vaswani17},
which we call the Mass-Conserving Transformer.
Due in part to the presence of radiographic noise,
the features identified by the radiograph-to-features
network can only be determined up to a certain level of 
precision.
The two networks have different approaches for 
handling this uncertainty.
The generative approach builds uncertainty into the 
model and is trained to minimize degeneracy 
of the density reconstructions.
The deterministic network uses mass-conservation to 
improve robustness of predictions.

Both architectures are trained on the features-to-density
problem using data at 
time steps $n=25,30,35,40$.
The training set consists of 80\% 
($N=$ 23,040) randomly selected 
density time series for each inner surface perturbation
profile and the testing set consists of the remaining
20\% ($N=$ 5,760) of data.
Since most of the action is downstream of the outgoing shock, we
primarily restrict our attention to reconstructing 
a smaller region encapsulating the gas metal interface. 
For purposes of comparison, the 
architectures were trained to output a 150$\times$150 square
pixel image representing the density field
in the cylindrical domain,
$\left[0, \frac{15}{44} L\right] \times \left[0, \frac{15}{44} L\right].$
This smaller domain window captures the most interesting 
physics. 
Additionally, the Mass-Conserving Transformer architecture
was trained to output a 440$\times$440 square pixel image 
representing the full domain of the simulation data, 
$[0, L] \times [0, L]$.

\subsection{ShockDecoderViT: A Vision Transformer-based Shock Decoder}

We propose a new generative
architecture to decode the density field from the
shock features based on the vision transformer~\cite{dosovitskiy2021an}
and a conditional variational autoencoder (cVAE) learning framework~\cite{kingma2013auto,cinelli2021variational,kingma2014semi}. 
In particular, the
data consists of the density field $\rho(t)$ and a corresponding small number of
numeric features characterizing the shock ${\cal F}(t)$ at four times in a
dynamically evolving flow. The aim is to obtain a trained model (in
this case just the decoder part of the cVAE) that,
given a set of time evolving shock features alone---the
conditions,
generates a sequence of 2D density
fields at the corresponding times.
We call the decoder a ShockDecoderViT
because the details of the flow can be reconstructed given
a minimal set of shock characteristics.
%We remind the reader, however, that
In common with most current ML approaches, the utility of this
model is largely restricted to the training data distribution.

The cVAE consists of an encoder and a decoder. First, the encoder
embeds the input images and conditions and processes them using the
transformer. Here ideas of the vision transformer are used to process
the density field. That is, the density field is considered as an
image that is first split into patches which serve as tokens.  In the
present setting, 
%approximately 
only the region interior of the
outermost extent of the outgoing shock is considered for the reason
that most of the rapid and large dynamical changes are confined to
this region. In particular, the interior most $150 \times 150$ region is
partitioned into 10 patches of size $15 \times 15$.
The patches and the conditions, which are taken to be 
the zeroth through seventh cosine harmonics of the outgoing shock
at the four times, are then projected onto the tokens' embedding space. 
After adding positional encoding to the projections, 
%tokens (read, their embedding) 
% the tokens are processed using the popular 
dot-product attention-based transformer blocks  
are applied
to obtain a
compact variational latent space representation of the temporal
sequence of densities and shock features at the bottleneck.  That is, the encoder
produces the mean $\mu(\rho(t), {\cal F}(t))$ and variance
$\Sigma(\rho(t), {\cal F}(t))$ of the latent space features as functions of density and shock features.  
Thereafter, the decoder takes inputs of
the latent variables realizations 
and the shock features and embeds them into a series 
of tokens
that are processed using attention-based transformer blocks 
before they are projected back into density space.

The transformer-based encoder and decoder (ShockDecoderViT) of the
cVAE are trained simultaneously with the Evidence Lower Bound (ELBo)
of variational inference as the loss function.
This loss function may be thought of as having two components:
the reconstruction loss of the autoencoder
and the KL divergence between the latent distribution produced by the encoder
and a standard multivariate normal.
In the testing phase, only the decoder is used;
it takes as input the shock features.
%As mentioned earlier,
The shock features
% These conditions 
are combined with random realizations of the latent space
variables to then generate the density fields.

In summary, our new architecture, the ShockDecoderViT, combines the
strengths of the dot product attention-based Vision Transformer and
the conditional variational autoencoder to generate temporally
coherent sequences of density fields from dynamically evolving
sequences of shock features. While the patch-based approach allows it
to handle large images efficiently, the transformer provides the
ability to capture long range dependencies in both the spatial and
temporal domains.  Results are presented for an architecture that uses
two transformer blocks in the encoder and eight transformer
blocks for the decoder. Increasing the number of encoder transformer
blocks and small variations of the number of transformer blocks in the
decoder led to minor changes in results (not shown).

% end remove

\subsection{Mass-Conserving Transformer Architecture}

In addition to considering the ShockDecoderViT, 
which is a generative network, 
we also investigated a purely deterministic architecture. 
The Mass-Conserving Transformer network is 
a structure-preserving architecture 
that uses the transformer blocks introduced in~\cite{Vaswani17}. 
While the ShockDecoderViT uses a vision transformer in 
both the encoder and decoder to go between 
latent representations and density field image patches,
the Mass-Conserving Transformer treats the input shock features using 
a series of transformer blocks before projecting to density.
The network uses a mass-conservation layer
to enforce that the dynamics conserve mass between time steps.

The input to the Mass-Conserving Transformer is
a matrix $x\in\mathbb{R}^{N_t\times N_f}$,
and the output is $y\in\mathbb{R}^{N_t\times N_r \times N_z}$.
$N_t$ is the number of times in the sequence, $N_f$
is the total number of features, which consists of shock features
a temporal encoding vector which represents polynomials in $n$,
$n^i$, for $i=0, 1, \dots N_p$,
and $N_r\times N_z$ is the resolution of the output image.
\remove{Consider a number of known shock and edge features 
as a function of time step, $n=1, \dots, N_t$, where $N_t$ is the number of times in the temporal sequence.
Define the
shock features $a_m^{n}$, $m=1,\dots, N_s$,
edge features $b_m^{n}$, $m=1,\dots, N_e$,
and temporal encoding features $c_m^{n}$, $m=1,\dots, N_p$,
where $N_s, N_e, N_p$ are the number of shock, edge, and temporal encoding features used, respectively.
The inputs to the network are the matrix
\begin{align*}
x = 
\left[\begin{array}{ccc|ccc|ccc}
     a_1^1 & \dots & a_{N_s}^1 &
     b_1^1 & \dots & b_{N_e}^1 & 
     c_1^1 & \dots & c_{N_p}^1 \\
     a_1^2 & \dots & a_{N_s}^2 &
     b_1^2 & \dots & b_{N_e}^2 & 
     c_1^2 & \dots & c_{N_p}^2 \\
     \vdots & \vdots & \vdots &
     \vdots & \vdots & \vdots & 
     \vdots & \vdots & \vdots \\
     a_1^{N_t} & \dots & a_{N_s}^{N_t} & 
     b_1^{N_t} & \dots & b_{N_e}^{N_t} & 
     c_1^{N_t} & \dots & c_{N_p}^{N_t}\\
\end{array}\right]\in\mathbb{R}^{N_t\times N_f},
\end{align*}
where $N_f = N_s+N_e+N_p$ is the total number of features.
The temporal encoding features are used to distinguish 
ordering in the input sequence. These features are 
chosen to be polynomials in $n$.
\begin{align}
    c_i^n = n^i, \qquad i=1,\dots N_p, \quad n = 1,\dots, N_t.
\end{align}}
The shock features are each centered and normalized using the mean and standard deviation of the training set. 

The features are first inputted into a network consisting of multiple transformer blocks,
\begin{align}
    \chi = T_B \circ \dots \circ T_1 (x),
\end{align}
where $\chi\in\mathbb{R}^{N_t\times N_f}$ is a matrix 
representing a latent representation of the density fields at each
time in the sequence.
The transformer blocks have the same structure as the blocks
proposed in~\cite{Vaswani17}.
Each block applies dot-product attention to incorporate temporal
dependencies.

After the series of transformer blocks, the latent variables are
projected into density fields.
A critical component of projection step is the learning of 
characteristic eigenfunctions,
$\mathcal{Z}^{(i)}\in\mathbb{R}^{N_r\times N_z}$, 
for $i=1, \dots, N_{\mathcal{Z}}$ on the image plane.
The first set of transformations is given by
\begin{align}
    \mathcal{Y}_{ij}^n = h\left(
    \mathcal{Z}_{ij}^{(1)}, \dots, \mathcal{Z}_{ij}^{(N_{\mathcal{Z}})};\;\theta_n\right),
    \qquad i = 1,\dots, N_r,\quad j= 1,\dots, N_z, \quad n=1,\dots, N_t,
\end{align}
where $h$ is a feedforward neural network parameterized by weights,
$\theta_n$, which are in turn determined by
\begin{align}
    \theta_n = g(\chi_n), \qquad n=1,\dots, N_t,
\end{align}
where $g$ is also a feedforward neural network.
The intermediate output, 
$\mathcal{Y}\in\mathbb{R}^{N_t\times N_r \times N_z}$,
represents a sequence of images with the same resolution at 
the final output $y$.
\remove{The $j^{\rm th}$ feature of $\chi$ is treated as a coefficient 
in the linear expansion
%for a corresponding learned image $\rho_j$ in the sum
\begin{align}
    \bar{y}^n
    = 
    \sum_{j=1}^{N_f}\chi_{nj} \mathcal{Y}_j, \qquad n=1,\dots,N_t,
\end{align}
where $\chi_{nj}$ is the $(n, j)$ matrix 
component of $\chi$, and $\mathcal{Y}_j$ and $\bar{y}^n$ both have the 
dimensions of the output density image.}
The reconstructed density is obtained by passing 
the sequence, $\mathcal{Y}^n$, $n=1,\dots,N_t$, 
through a mass conservation layer. 
Let $\mathcal{M}$ be the total mass of the object and let 
\begin{align}
M(\gamma) = \sum_{i, j} 4 \pi r_i \gamma_{ij} \Delta r \Delta z,
\end{align}
be the calculation of mass of in the image $\gamma$, where $r_i$ is the radius at index $i$ and $\gamma_{ij}$ 
is the density value in pixel $(i, j)$.
The output density reconstruction is the sequence of images 
\begin{align}
    y^{n} = \frac{\mathcal{M}}{M( |\mathcal{Y}^{n}|)} |\mathcal{Y}^{n}|, \qquad n=1,\dots, N_t.
\end{align}

Our architecture used 6 consecutive transformer blocks, $B=6$,
each with 8 heads, $H=8$,
a latent dimension of 64, $k=64$,
and a feedforward neural network with inner dimension 2048
and $\tanh$ activation function.
For the projection layer, we chose $N_{\mathcal{Z}}=40$ eigenfunctions
and represented both $g$ and $h$ using feedforward neural networks
with 2 hidden layers and latent dimension of 100.
The architecture was trained to minimize the root-mean-squared error (RMSE)
of the density reconstructions on the training set.

\section{Results}
\label{sec:results}

\remove{
\textbf{Proposed outline}:
\begin{itemize}
    \item Introduce an outline for the section. Describe any commonalities and differences 
    between the training of the three architectures.
    Describe the partitioning of the dataset into training, testing,
    and, if applicable, validation.
    Describe the time steps chosen for each method and the pixel 
    resolution used by each method.
    \item Figure: histogram of RSME evaluated on testing set for each method
    \item VAE Results
    \begin{itemize}
        \item Any VAE specific details. 
        \item Figure: reconstructions from VAE
        \item Figure: sampling of the latent space. Show the absence of degeneracy 
    \end{itemize}
    \item GAN Results
    \begin{itemize}
        \item Any GAN specific details.
        \item Figure: reconstructions from GAN
        \item Figure: reconstructions with using less harmonics in the training set.        
        \item Figure: training loss by using less harmonics in the input features. Show that taking away harmonics results in a higher loss.
    \end{itemize}
    \item Transformer
    \begin{itemize}
        \item Any transformer specific details
        \item Figure: reconstructions from transformer
        \item Figure: Full domain reconstructions - zoom in on RMI
        \item Figure: show corruptions due to noise from Xiaojian's feature extractor
        \item Figure: Capturing peak and trough in RMI and comparing with 
        ground truth
    \end{itemize}
\end{itemize}
}

This section presents results of the trained 
ShockDecoderViT and Mass-Conserving Transformer
networks.
First we compare the two approaches using 
ensemble metrics such as root-mean-squared error (RMSE) and 
structural similarity.
Section~\ref{sec:shockdecoderresults} examines
the generative nature of ShockDecoderViT.
Section~\ref{sec:masconsresults}
analyzes the density reconstructions of 
each inner surface perturbation profile
and its corruptions due to errors arising from the 
radiograph-to-features network.
Additionally, we use an example reconstruction 
to demonstrate how peak-to-trough evolution can be 
calculated accurately, as indicated by capturing the RMI growth rates from the reconstruction
and comparing to those obtained from the ground truth simulations.

\begin{figure}[htb]
  \centering
  \includegraphics[trim=0 0 0 0, clip,
  width=.49\textwidth]{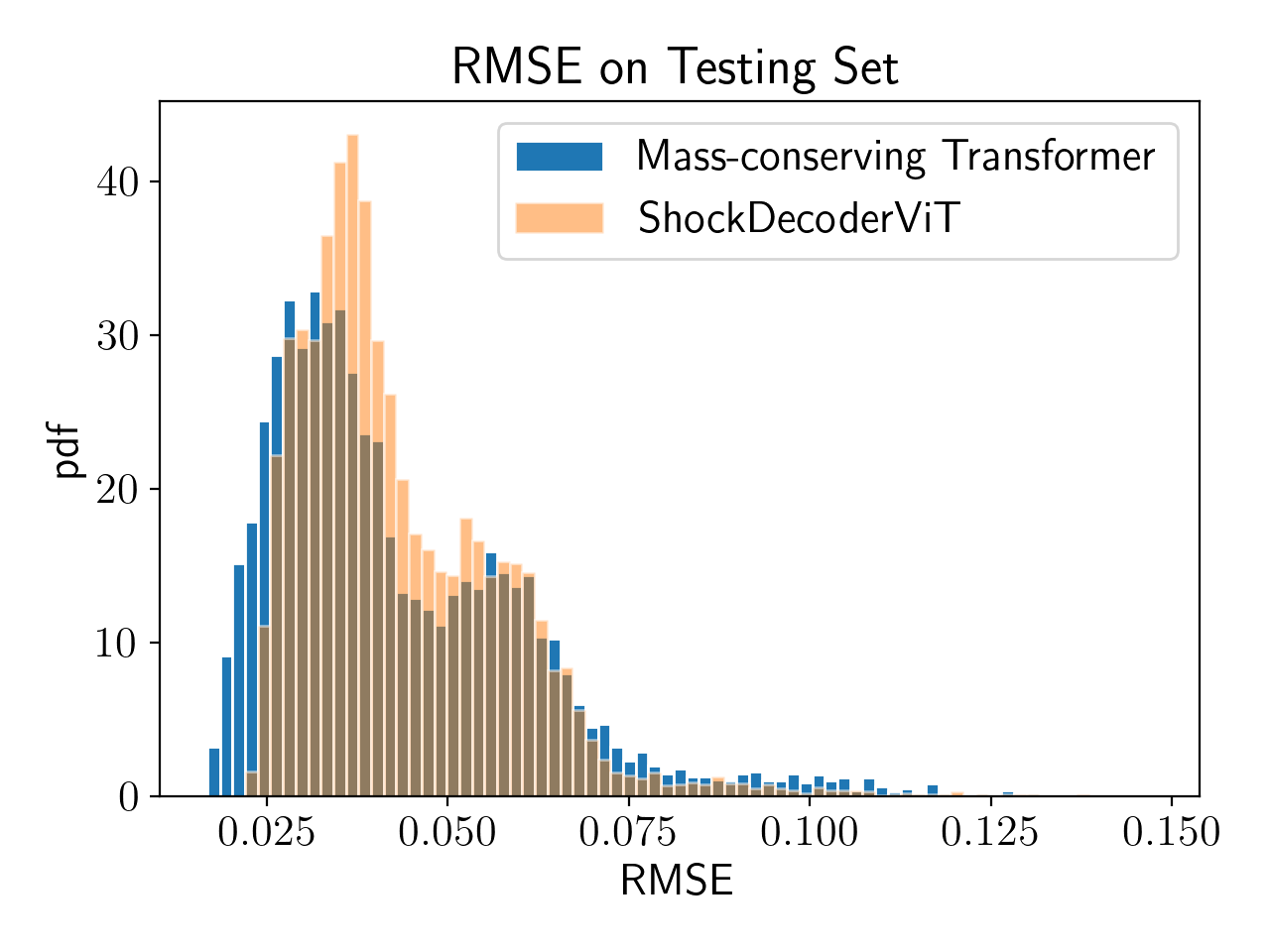}  
  \includegraphics[trim=0 0 0 0, clip,
  width=.49\textwidth]{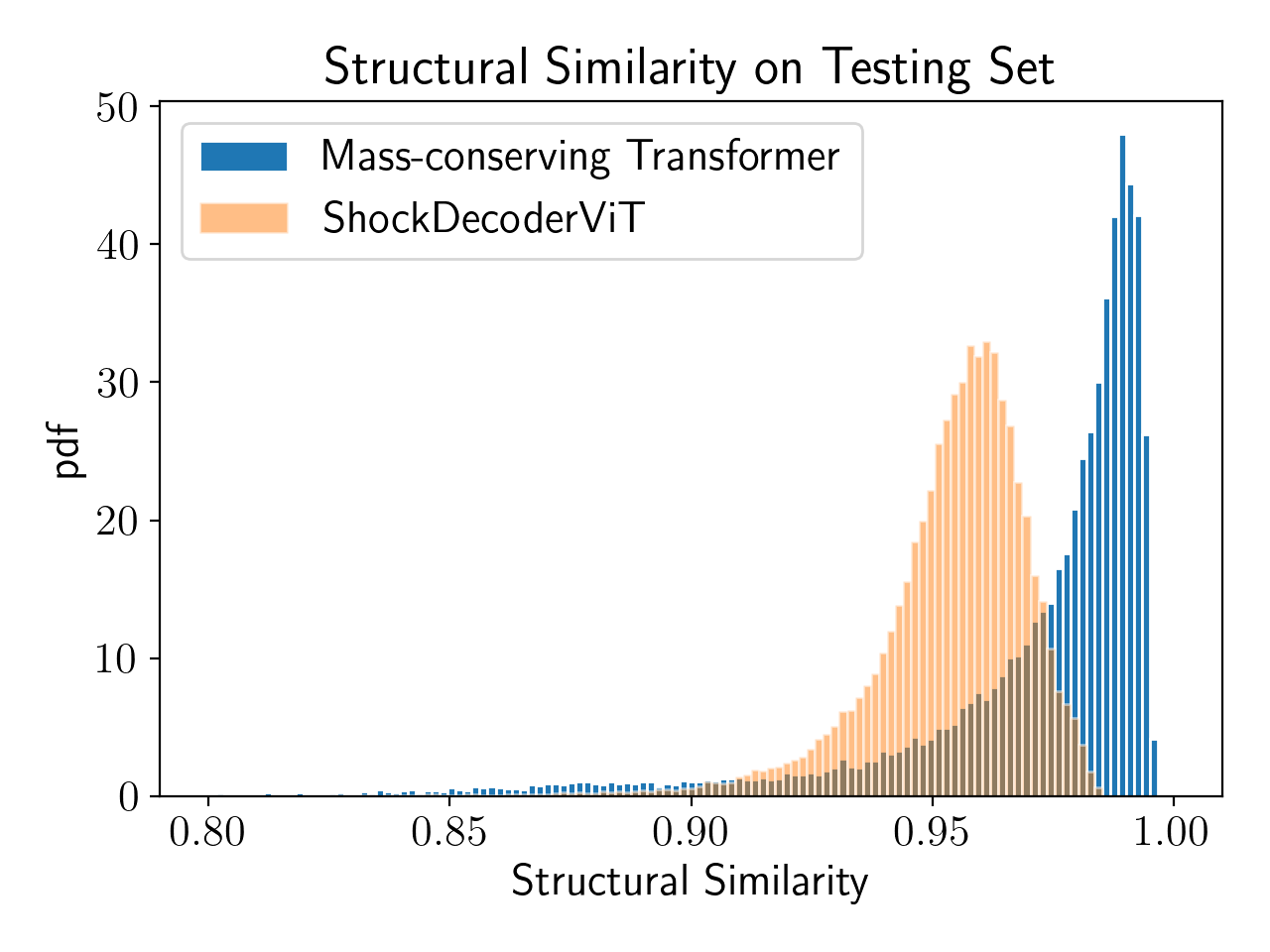} 
  \begin{tabular}{|c|c|c|c|c|c|c|}
  \hline
  \multirow{2}{*}{Architecture} 
  & \multicolumn{3}{c|}{RMSE (g/cc)}
  & \multicolumn{3}{c|}{Structural Similarity} \\
  \cline{2-7}
   & min & average & max & min & average & max\\
  \hline
ShockDecoderViT & 0.023 & 0.044 & 0.148 & 0.58 & 0.95 & 0.98 \\
%ShockDecoderViT & 0.048 & 0.092 & 0.312 & 0.58 & 0.95 & 0.98 \\ %0.79 & 1.52 & 5.19 & 0.58 & 0.95 & 0.98 \\
\hline
Mass Conserving Transformer & 0.017 & 0.043 & 0.147 & 0.76 & 0.97 & 1.00 \\
%Mass-Conserving Transformer & 0.035 & 0.091 & 0.310 & 0.76 & 0.97 & 1.00 \\%0.59 & 1.51 & 5.16 & 0.76 & 0.97 & 1.00 \\
\hline
  \end{tabular}
  \caption{Histogram of root-mean-squared errors (RMSE) (top left) and structural similarity (top right) between the
  density reconstruction and ground truth for the testing set
  for the ShockDecoderViT and Mass-conserving transformer.
  Bottom: table of summary statistics for the above histograms.
  }
  \label{fig:histcomparison}
\end{figure} 

Figure~\ref{fig:histcomparison} shows histograms of 
RMSE and structural similarity evaluated on the testing set for 
both architectures along with a table of summary statistics. 
Both networks exhibit similar performance in RMSE, with 
the Mass-Conserving Transformer having better minimum, 
average, and maximum errors compared to the ShockDecoderViT.
The Mass-Conserving Transformer significantly outperforms the
ShockDecoderViT in structural similarity, with a mode shifted to 
the right and better minimum, average, and maximum structural similarity values.

\subsection{ShockDecoderViT Density Reconstruction}
\label{sec:shockdecoderresults}

\begin{figure}[htb]
  \centering
  \includegraphics[trim=0 0 0 0, clip,
  width=.49\textwidth]{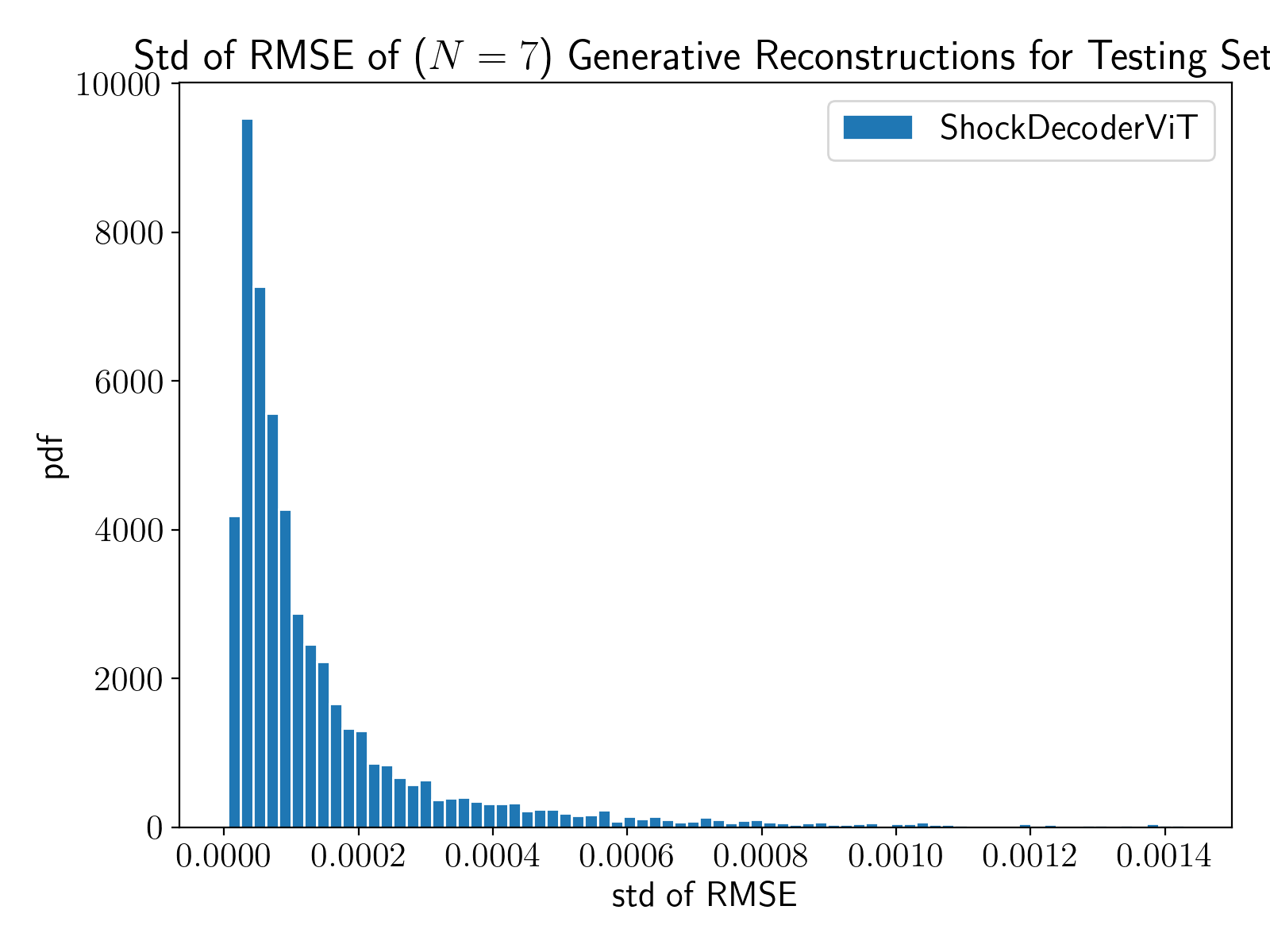}  
  \includegraphics[trim=0 0 0 0, clip,
  width=.49\textwidth]{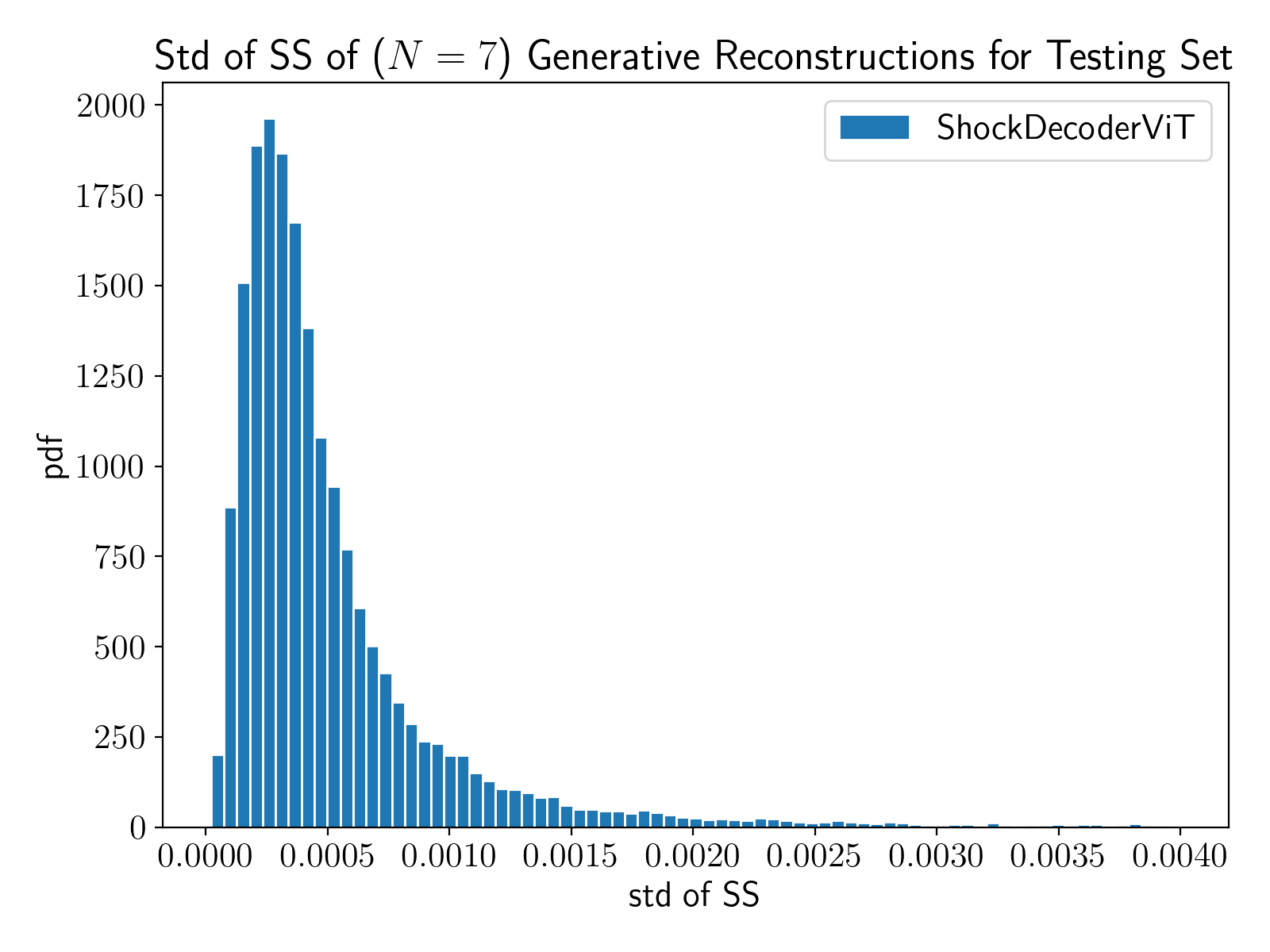}
  \resizebox{\columnwidth}{!}{
  \begin{tabular}{|c|c|c|c|c|c|c|}
  \hline
  \multirow{2}{*}{Architecture} 
  & \multicolumn{3}{c|}{Std of RMSE $(N=7)$ (g/cc)}
  & \multicolumn{3}{c|}{Std of Structural Similarity $(N=7)$} \\
  \cline{2-7}
   & min & average & max & min & average & max\\
  \hline
ShockDecoderViT & 4.82e-06 & 2.56e-04 & 2.27e-02 & 2.09e-05 & 5.83e-04 & 4.14e-02 \\
% ShockDecoderViT & 1.01e-05 & 5.39e-04 & 4.78e-02 & 2.09e-05 & 5.83e-04 & 4.14e-02 \\ %1.69e-04 & 8.97e-03 & 7.96e-01 & 2.09e-05 & 5.83e-04 & 4.14e-02 \\
\hline
Mass-Conserving Transformer & 0 & 0 & 0 & 0 & 0 & 0 \\
\hline
  \end{tabular} }
  \caption{Histogram of the standard deviation of RMSE (left) and SS (right) for $N=7$ generations of each example in the testing set. 
  Bottom: table of summary statistics for the above histograms.}
  \label{fig:stdhist}
\end{figure}

This section examines the generative nature of the 
ShockDecoderViT network.
For each set of shock features in the testing set, we generated
$N=7$ realizations of density reconstructions using the 
ShockDecoderViT and computed the standard deviation of
the density, RSME, and structural similarity.
Figure~\ref{fig:stdhist} summarizes the results of this study,
showing histograms of the standard deviations of both RMSE
and structural similarity, along with summary statistics in a table.
The variation in the RMSE and Similarity metric are very small relative to the ensemble averages,
indicating that the variations due to the degeneracy are small.
We reiterate that the standard deviation of deterministic methods,
including the Mass-Conserving Transformer,
on a fixed input
is zero by definition.

\begin{figure}[htb]
  \centering
      \begin{subfigure}{\linewidth}
        \centering
  \includegraphics[trim=3cm 1cm 0cm 0, clip,
  width=.7\textwidth]{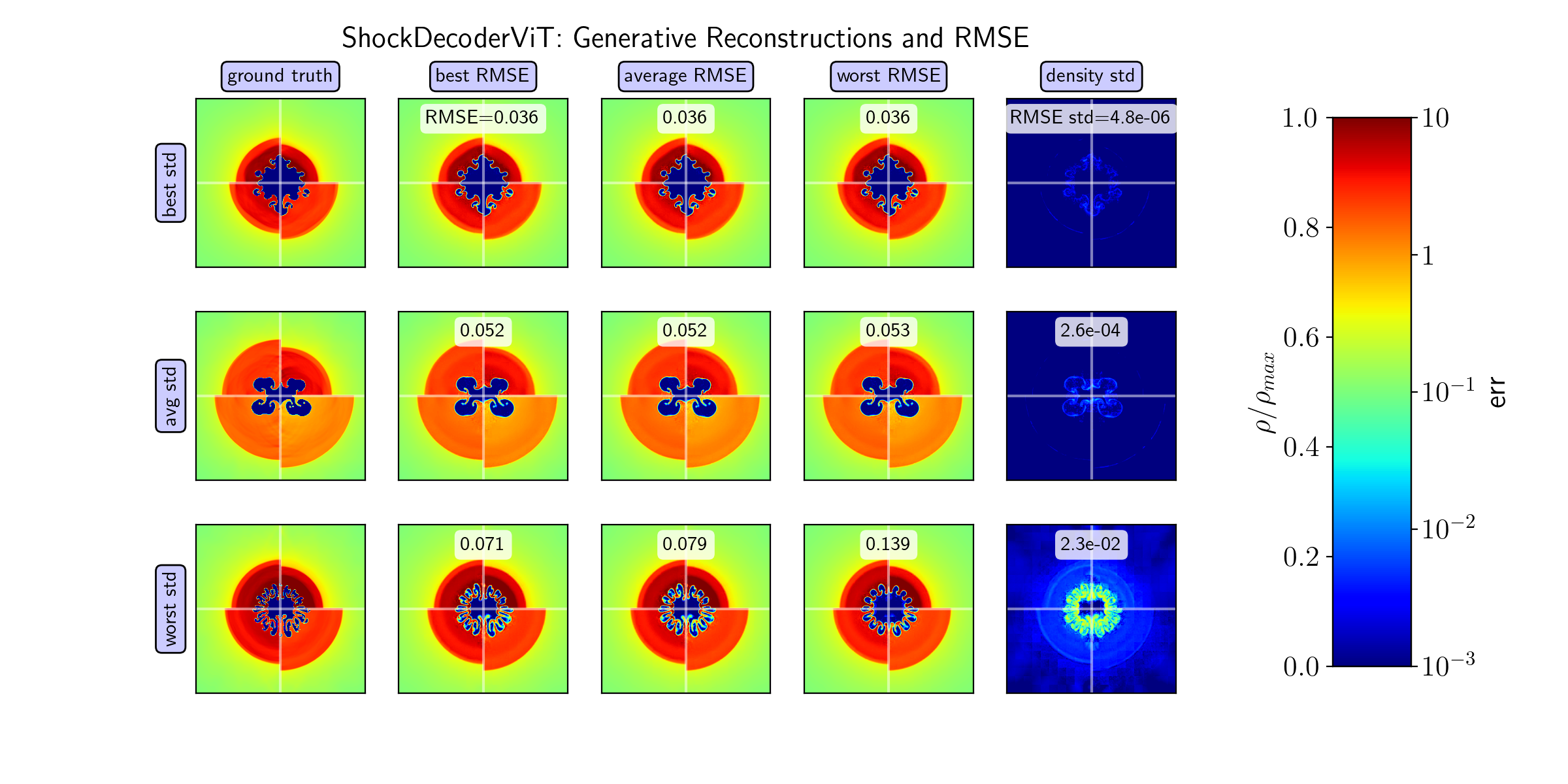}
        \subcaption{Multiple generative reconstructions for examples with the lowest (top row), average (middle row), and highest (bottom row) standard deviation in RMSE.}
    \end{subfigure}
    \vfill
    \begin{subfigure}{\linewidth}
        \centering
  \includegraphics[trim=3cm 1cm 0cm 0, clip,
  width=.7\textwidth]{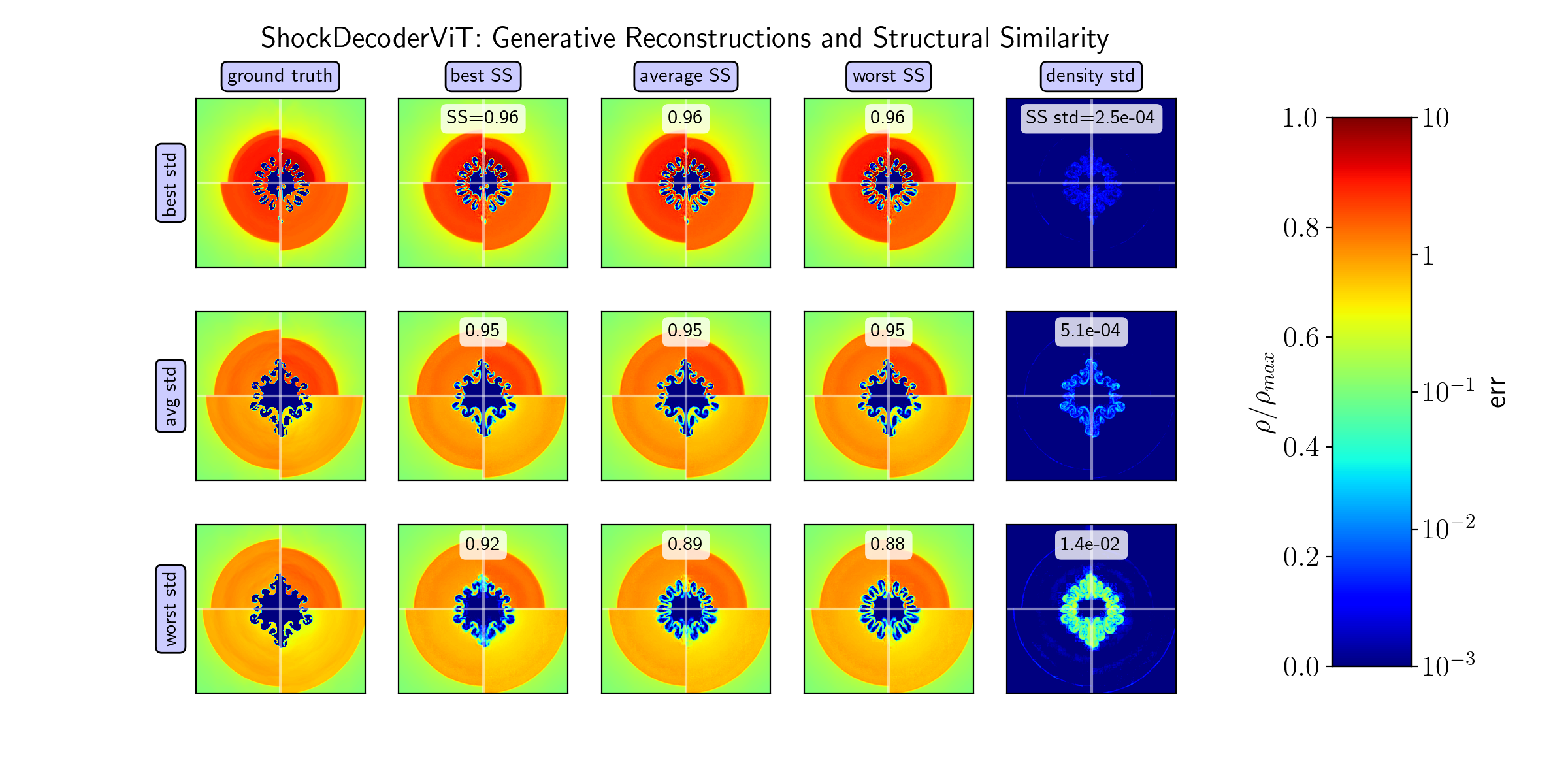}
         \subcaption{Multiple generative reconstructions for examples with the lowest (top row), average (middle row), and highest (bottom row) standard deviation in SS.}
    \end{subfigure}  
  \caption{Examples of generative reconstructions produced by the ShockDecoderViT corresponding 
  to the lowest, average, and highest 
  standard deviation in RMSE (a) and Structural Similarity (b). In both sub-figures, the
  left-most column is the ground truth, 
  the middle three columns correspond to the 
  best, average, and worst errors, and the right-most
  column shows the standard deviation in density 
  for the $N=7$ generative reconstructions for the corresponding case.
  Each plot shows quarter planes corresponding data at each of the
  four times in the sequence counter-clockwise from the first quadrant.
  }
  \label{fig:generative}
\end{figure}  

Figure~\ref{fig:generative} shows examples of generations from 
six sets of shock features. These six features were chosen to correspond
to the extremities of the standard deviation histogram, 
minimums, averages, and maximums for the 
standard deviation of RMSE and structure similarity.
For each set of features, we chose the best, average, and 
worst performing reconstruction to display. Additionally, 
the standard deviation of density is shown in the last column.
For the best and average examples, the reconstructions 
are nearly identical under visual inspection, while 
the worst performing reconstruction exhibits significant 
errors from the ground truth and are visually different from the
other reconstructions in the example. 
This is also confirmed by examining the large values present in the
density standard deviation heat map.
This extreme example is however one example of a number of outliers
represented by the narrow wide tail of the histogram in Figure~\ref{fig:stdhist}.

\subsection{Mass-Conserving Transformer Density Reconstruction}
\label{sec:masconsresults}

\begin{figure}[htb]
  \centering
  \includegraphics[trim=0 0 0 0, clip,
  width=.45\textwidth]{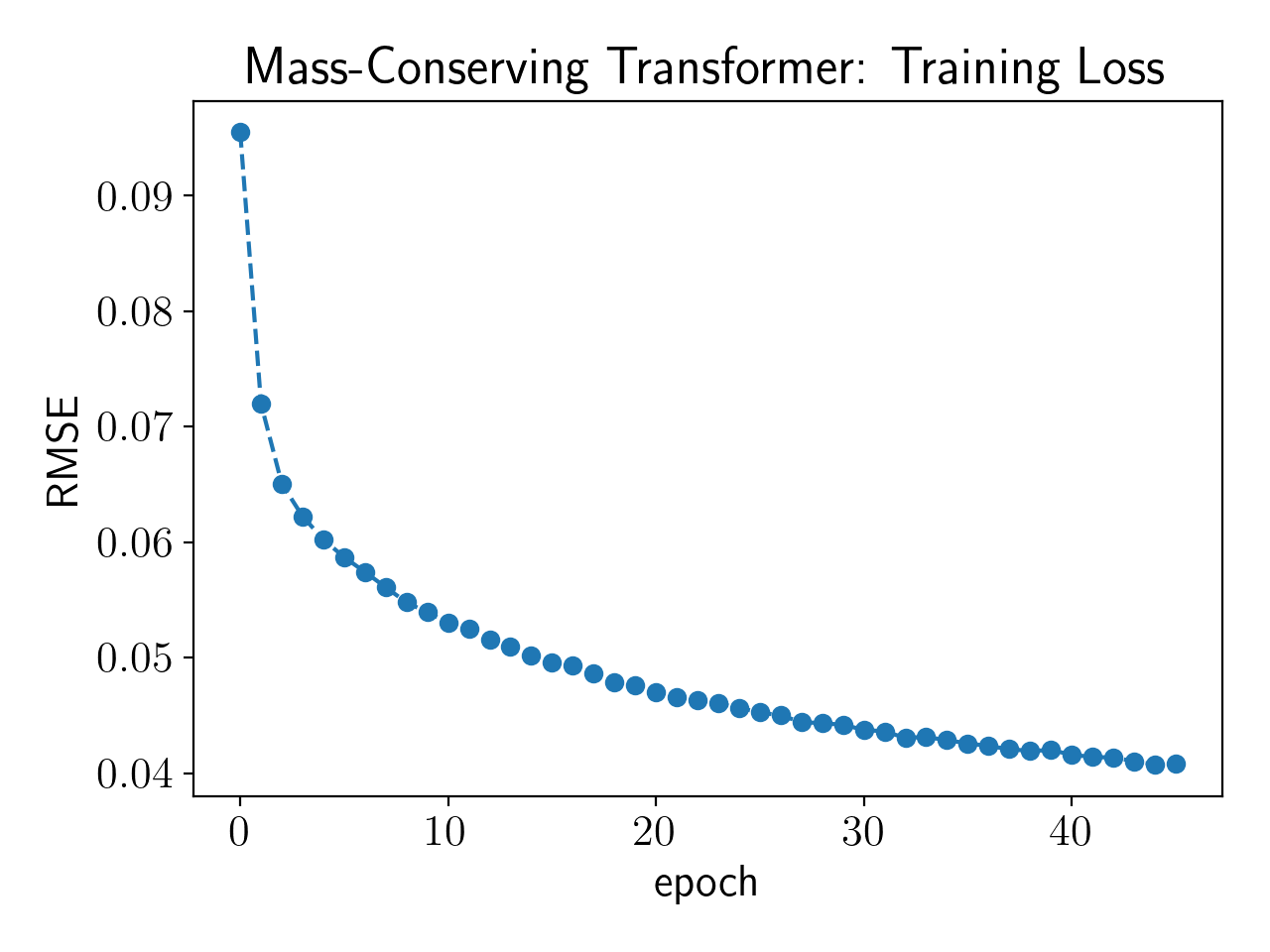}
  \caption{Root mean squared error
  verses epoch during training.
  }
  \label{fig:trainingerror}
\end{figure} 

\begin{figure}[htb]
  \centering
      \begin{subfigure}{\linewidth}
        \centering
  \includegraphics[trim=3cm 1cm 1.25cm 0, clip,
  width=.8\textwidth]{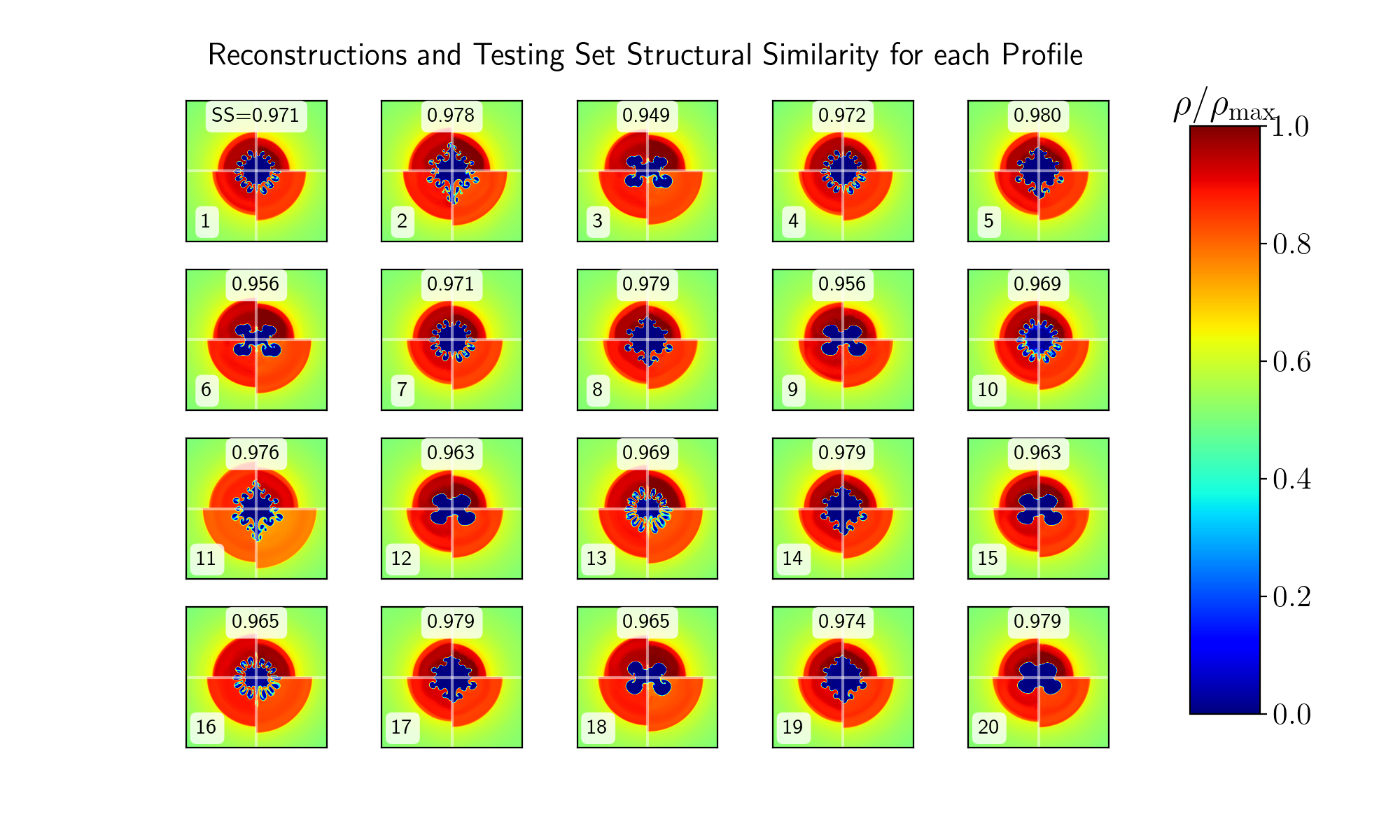}
        \subcaption{Mass-Conserving Transformer: Examples of density reconstructions for each inner surface perturbation profile sorted by the profile 
        ensemble average Structural Similarity in the testing set.
        The plots are labeled by their profile
        number and the corresponding profile ensemble average 
        Structural Similarity.}
    \end{subfigure}
    \vfill
    \begin{subfigure}{\linewidth}
        \centering
  \includegraphics[trim=3cm 1cm 1.25cm 0, clip,
  width=.8\textwidth]{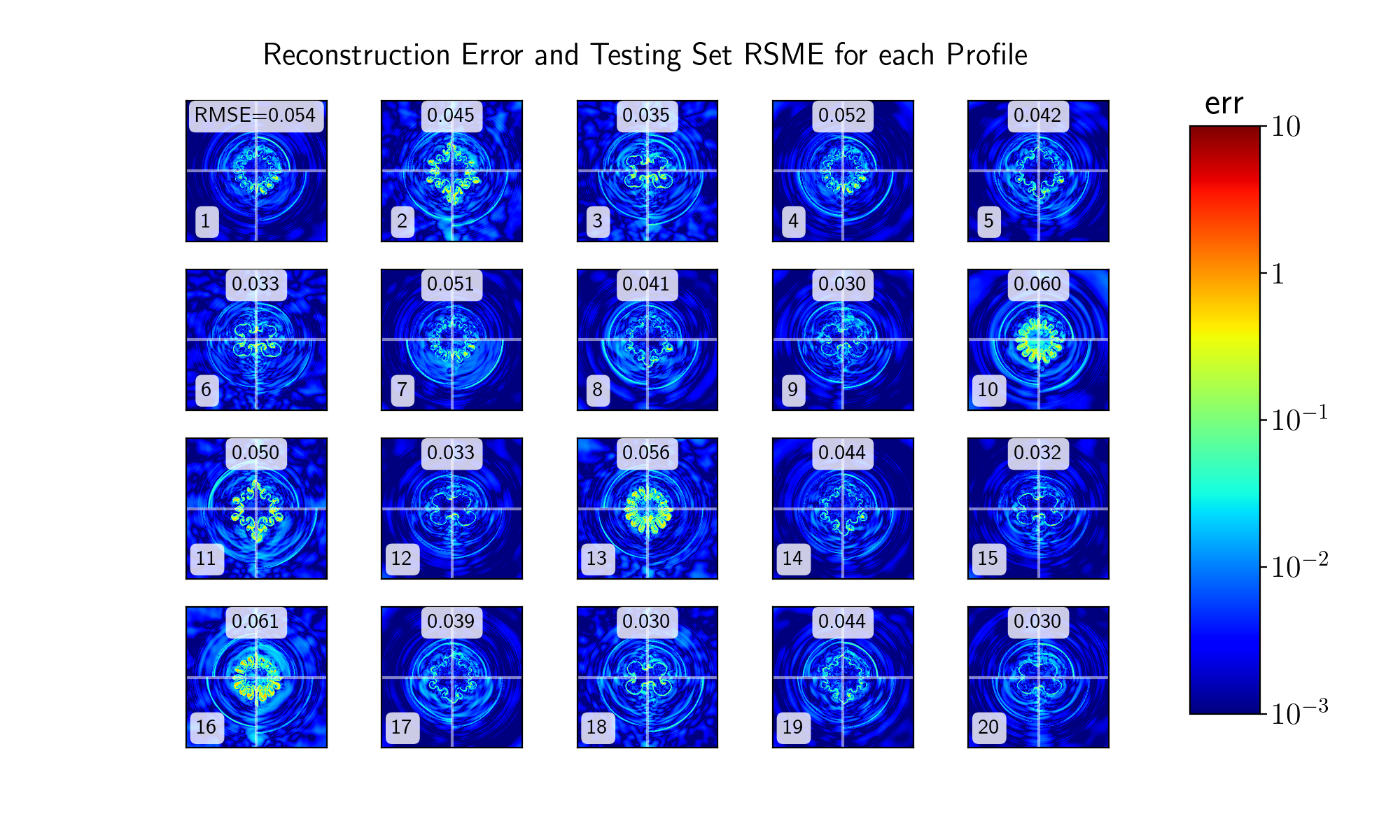}
         \subcaption{Mass-Conserving Transformer: Corresponding errors between the model and the ground truth on a log scale for the above reconstructions, labeled by their profile number
         and the corresponding profile ensemble average RSME in g/cc.}
    \end{subfigure}  
  \caption{Mass-Conserving Transformer:
  (a) Density reconstructions and (b) their errors 
  for a random choice of simulation parameters corresponding 
  to each inner surface perturbation profile.
  Each plot shows quarter planes corresponding data at each of the
  four times in the sequence counter-clockwise from the first quadrant.
  }
  \label{fig:recon_profiles}
\end{figure}

This section analyzes the density reconstructions 
produced by the Mass-Conserving Transformer network.
Figure~\ref{fig:trainingerror} shows the training loss,
representing RMSE over the training set,
verses the training epoch.
The trained model was used to produce the density reconstructions
and their corresponding errors shown in 
Figure~\ref{fig:recon_profiles}.
\remove{Each example reconstruction represents a random set of 
shock features for each inner-surface perturbation profile.}
Each example reconstruction in this figure, labeled 1 through 20, 
corresponds to a different set of shock features, chosen randomly,
corresponding to each inner-surface perturbation profile.
The labeled structural similarity and RMSE were computed for 
the ensemble of testing data for each profile. 
The top plot (a) displays the density reconstructions sorted 
from lowest to highest structural similarity.
The bottom plot shows the corresponding errors of the above plot,
labeled by profile ensemble average RMSE in g/cc.
Despite the low dimensional feature space, 
the density, including the complex details of the RMI, 
can be reconstructed to a high level of accuracy.

\begin{figure}[htb]
  \centering
  \includegraphics[trim=0 0 0 0, clip,
  width=.8\textwidth]{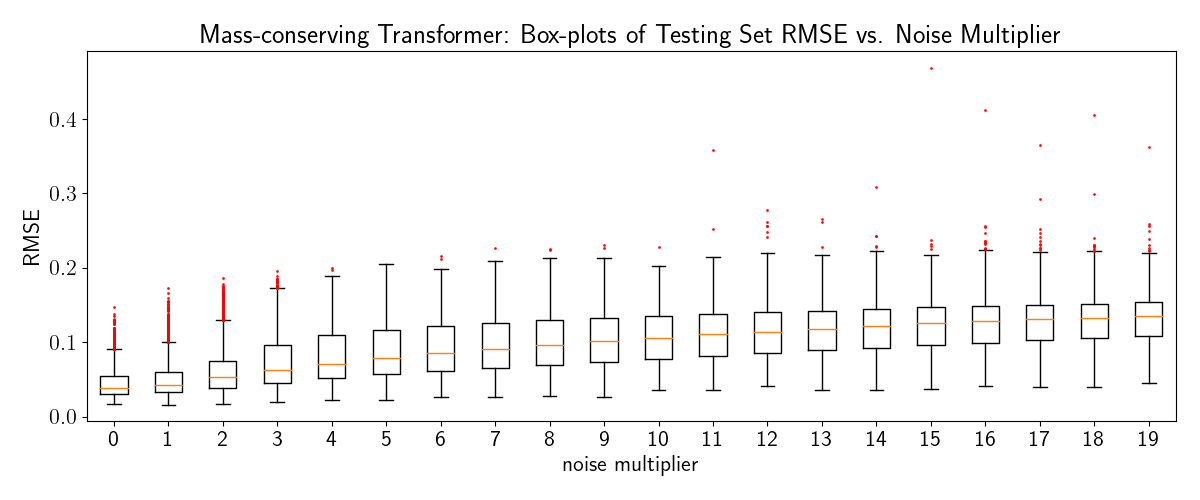}
  \caption{Box-plots of RSME (g/cc) evaluated on the testing 
  set corresponding to noise multipliers 
  $0, 1, \dots, 19$.
  Each box extends from the first quartile to the third quartile of the data with an orange line representing the median. The whiskers extend from the box to the farthest data point lying within 1.5x the inter-quartile range from the box. The red dots represent points outside of the whiskers. 
  }
  \label{fig:noise_boxplot}
\end{figure}

Next, we investigated the effect of errors from the radiograph-to-features 
network on the Mass-Conserving Transformer results.
For each set of shock features in the testing set, a random error
is generated using the noise model described in Section ~\ref{sec:noisemodel}.
Reconstructions were produced for this generation of random error
multiplied by factors of $0, 1, \dots, 19$, respectively,  and the corresponding RMSE was computed.
Figure~\ref{fig:noise_boxplot} shows box plots summarizing the 
error statistics of this study for each noise multiplier.
A clear degradation in accuracy is observed as the noise multiplier 
is increased. 
For the case of 1$\times$ multiplier, which corresponds
to the expected error produced from the radiograph-to-features network,
the boxplot represents a small shift from the 
clean features. 
The trend towards greater noise remains gradual and bounded, 
demonstrating the robustness of the model, which 
may be attributed to it's structure-preserving properties.
\remove{
Figure~\ref{fig:recon_noise} shows density reconstructions and
corresponding errors for multiple noise multipliers for
a shock feature set example chosen randomly.
For this chosen example, the reconstruction remains visually correct 
up until a multiplier of 7$\times$, after which it is observed that
the reconstruction morphs into an image resembling a different
plausible RMI profile.
As shown in the error plots, the location of the shock is still captured
accurately for all examples.
This demonstrates that, despite being in the presence of 
extreme out-of-sample noise, the network remains robust.}

\begin{figure}[htb]
  \centering
  \includegraphics[trim=0 0 0 0, clip,
  width=.45\textwidth]{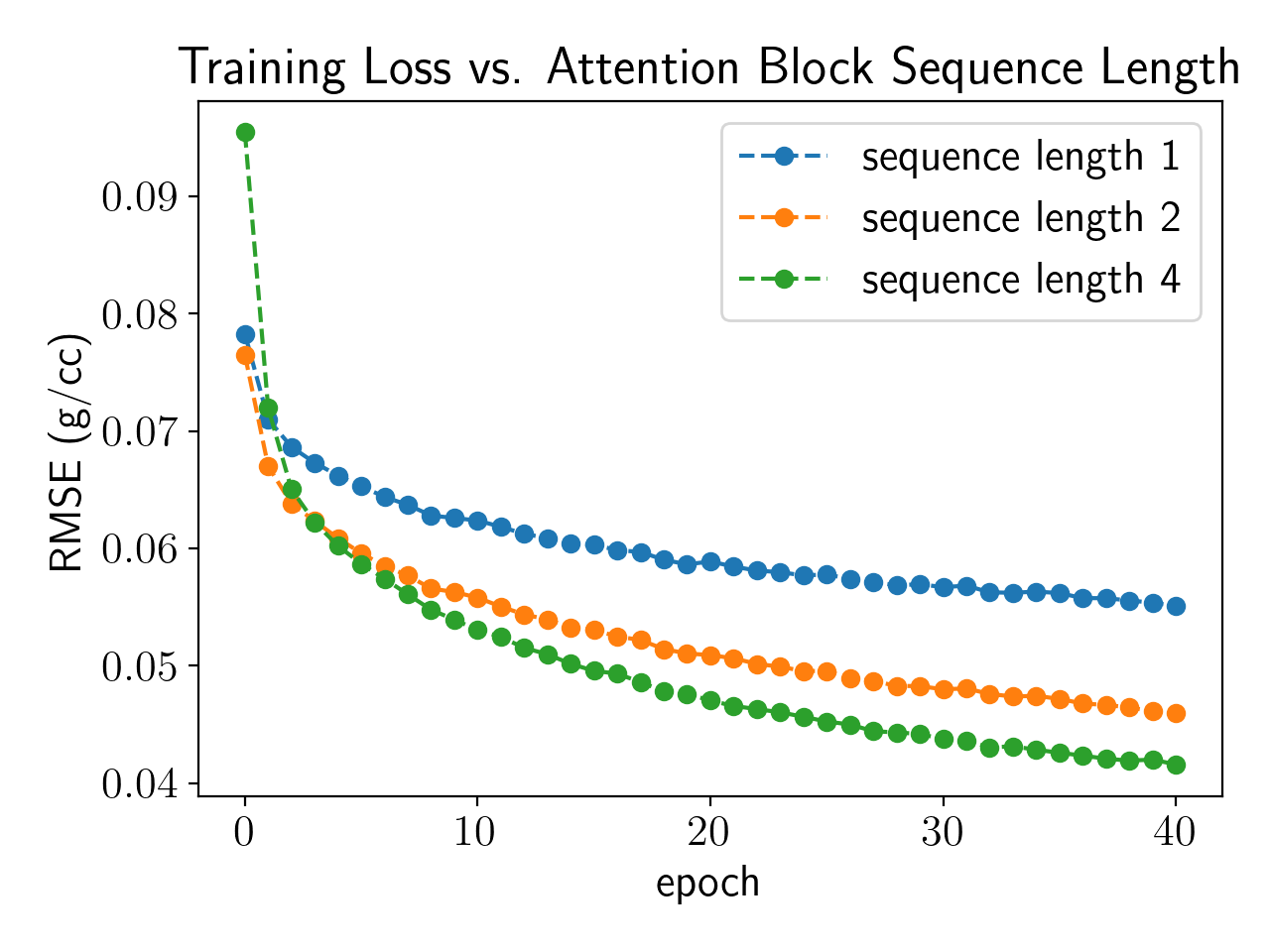}
  \caption{Comparison of training loss when the 
  attention blocks of the mass-conserving transformer 
  are fully connected (sequence length 4),
  connected only between the first two and latter two times
  (sequence length 2), 
  and fully disconnected (sequence length 1).
  For consistency between methods, RMSE is evaluated on the 
  entire sequence of 4 density reconstructions.
  }
  \label{fig:trainingerrorattn}
\end{figure} 

\remove{
\begin{figure}[htb]
  \centering
      \begin{subfigure}{\linewidth}
        \centering
  \includegraphics[trim=3cm 1cm 1.5cm 0, clip,
  width=.8\textwidth]{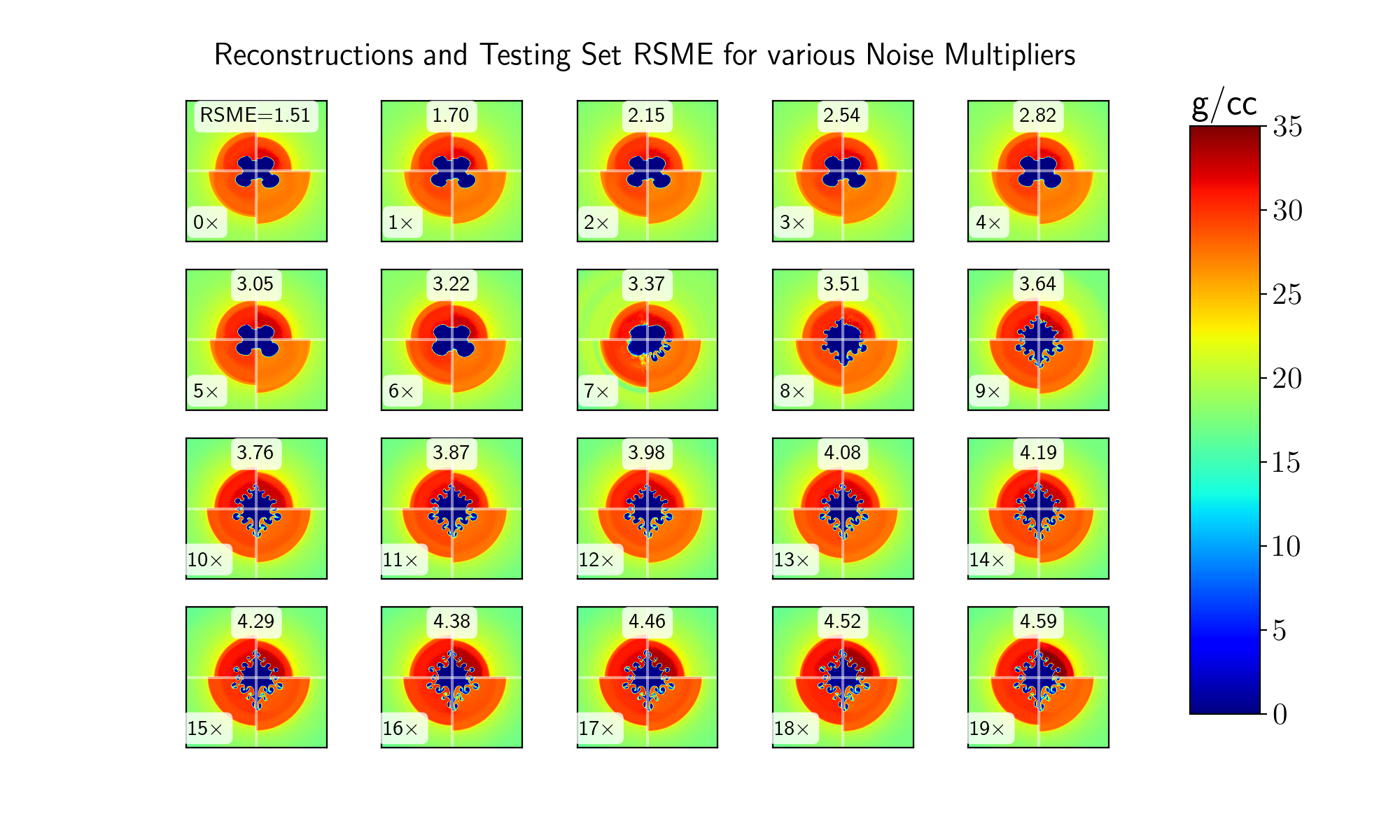}
        \subcaption{Mass-Conserving Transformer: Density reconstructions verses multiplicative factor on the noise.}
    \end{subfigure}
    \vfill
    \begin{subfigure}{\linewidth}
        \centering
  \includegraphics[trim=3cm 1cm 1.5cm 0, clip,
  width=.8\textwidth]{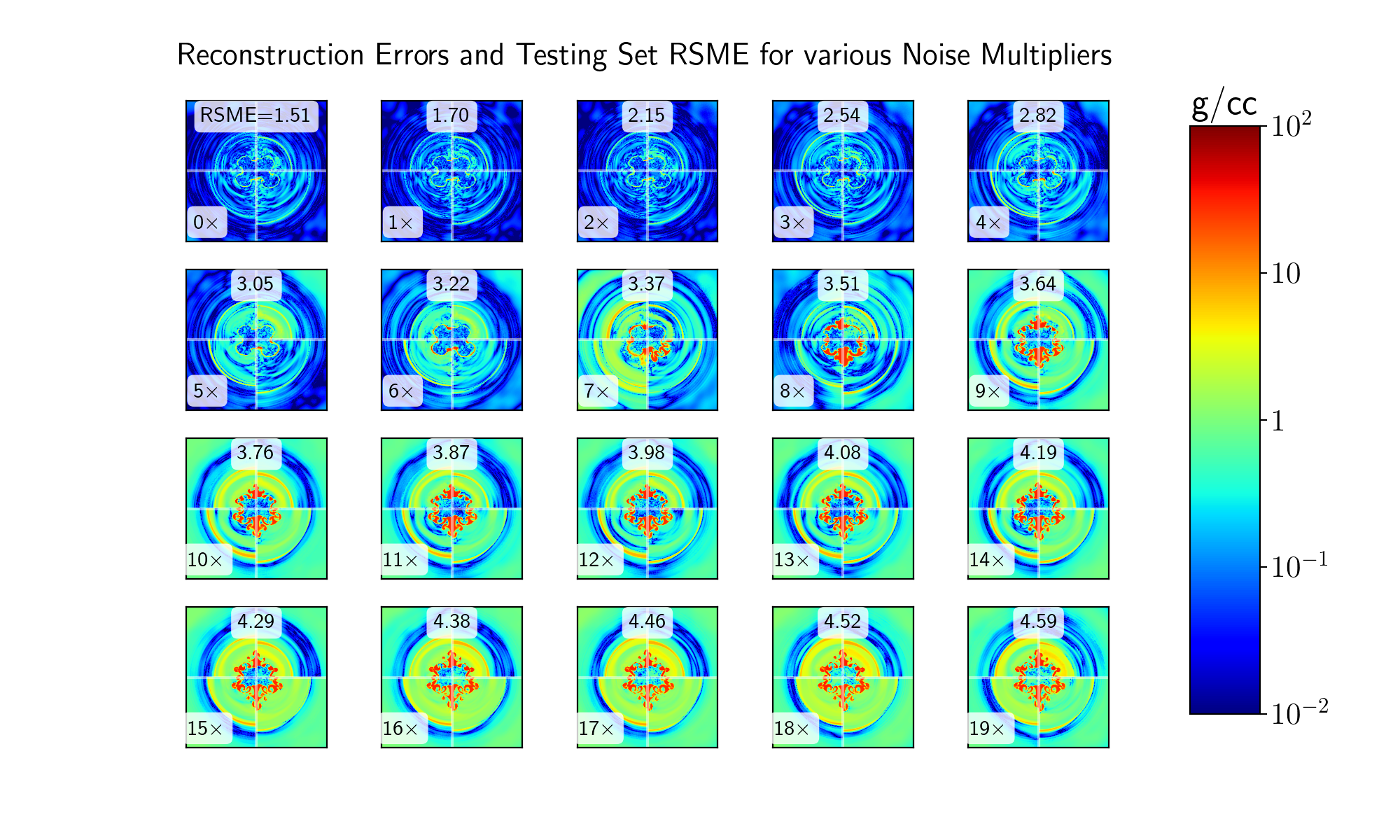}
         \subcaption{Mass-Conserving Transformer: Errors between the model and the ground truth on a log scale verses multiplicative factor on the noise.}
    \end{subfigure}  
  \caption{Mass-Conserving Transformer: Density reconstructions for a fixed feature-set that is gradually perturbed by a factor of single realization of noise from the feature extractor noise model. The multiplicative factor 0 in the top left 
  and is increased to 19 in the bottom right by steps of size 1. 
  Notably as noise is increased, the RMI profile begins to resemble the RMI profile corresponding to other 
  initial inner surface perturbation profiles.
  }
  \label{fig:recon_noise}
\end{figure}  
}

Next, we studied the expressive power of the 
attention mechanism used in the Mass-Conserving transformer.
By default, the dot product attention represents a
fully connected network between each of the 4 terms of the sequence.
For comparison, we also 
considered a sequence length 1 network, formed by breaking 
all the cross-temporal connections of the network,
and a sequence length 2 network, formed by breaking 
the cross-temporal connections between the first two times 
and the latter two times.
Figure~\ref{fig:trainingerrorattn} shows the training loss
of the three approaches. 
The results indicate that incorporating more temporal correlations 
led to smaller training losses.

\begin{figure}[htb]
  \centering
  \includegraphics[trim=0 0 0 0, clip,
  width=.49\textwidth]{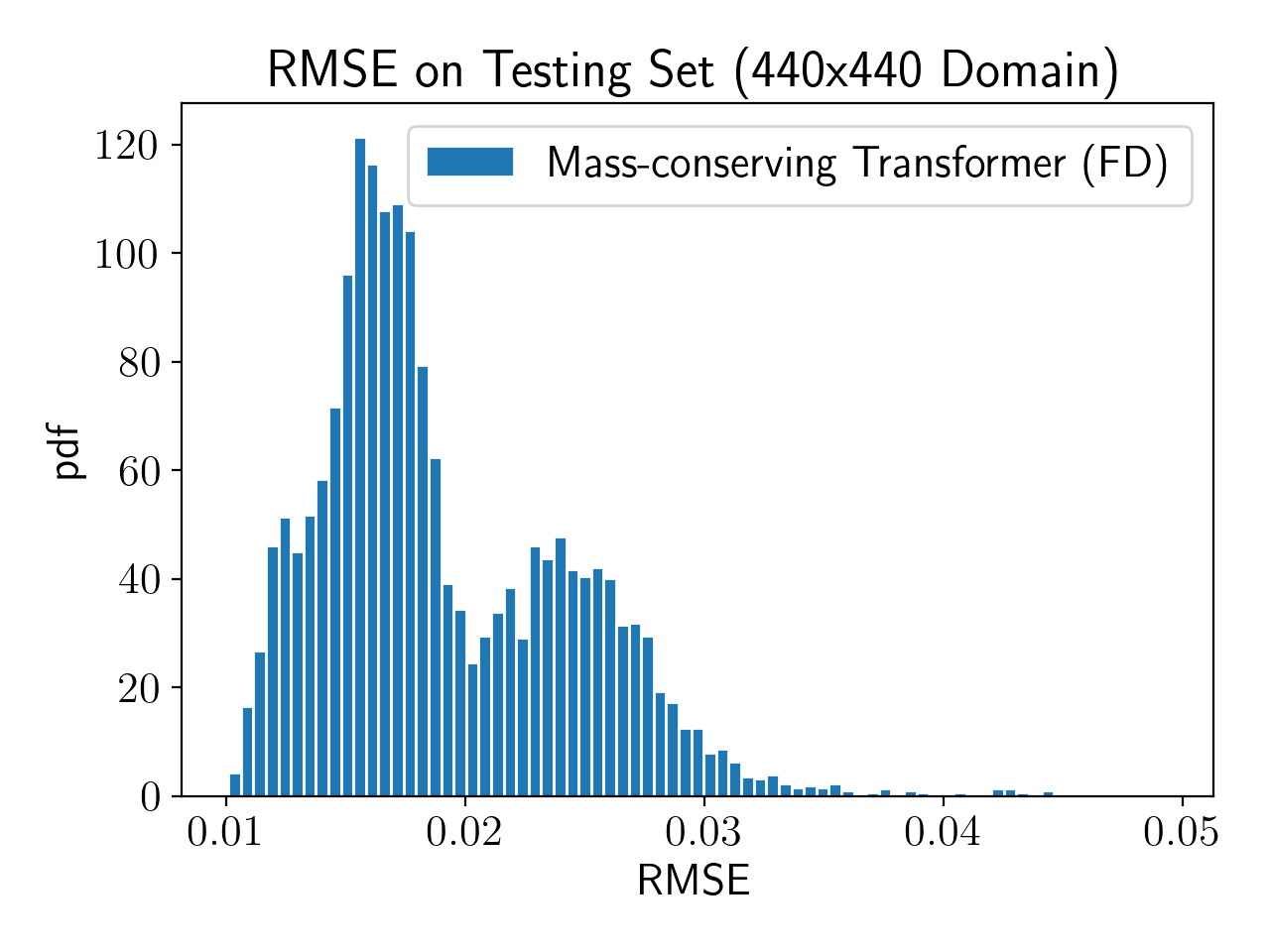}  
  \includegraphics[trim=0 0 0 0, clip,
  width=.49\textwidth]{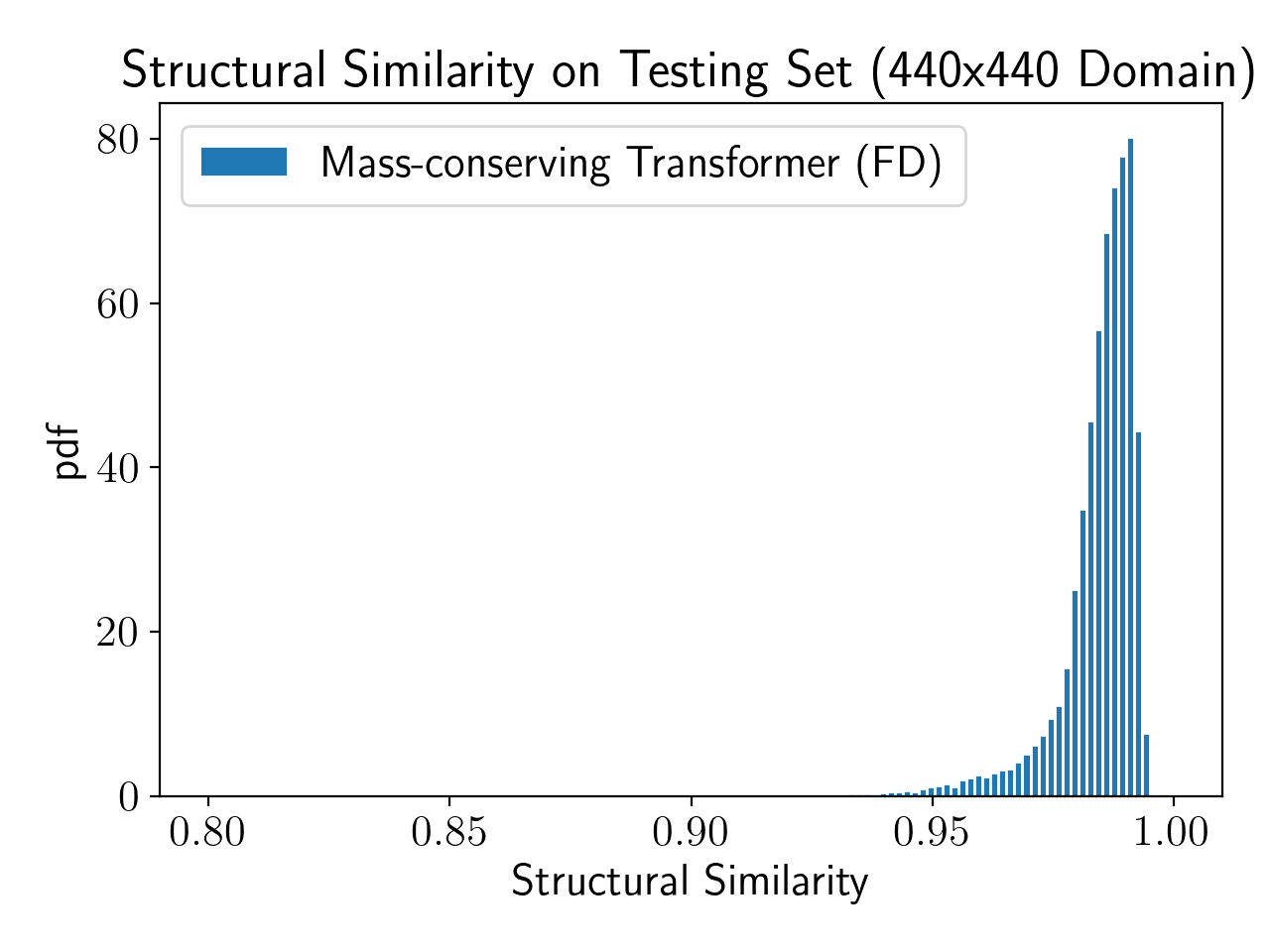} 
  \remove{\includegraphics[trim=0 0 0 0, clip,
  width=.49\textwidth]{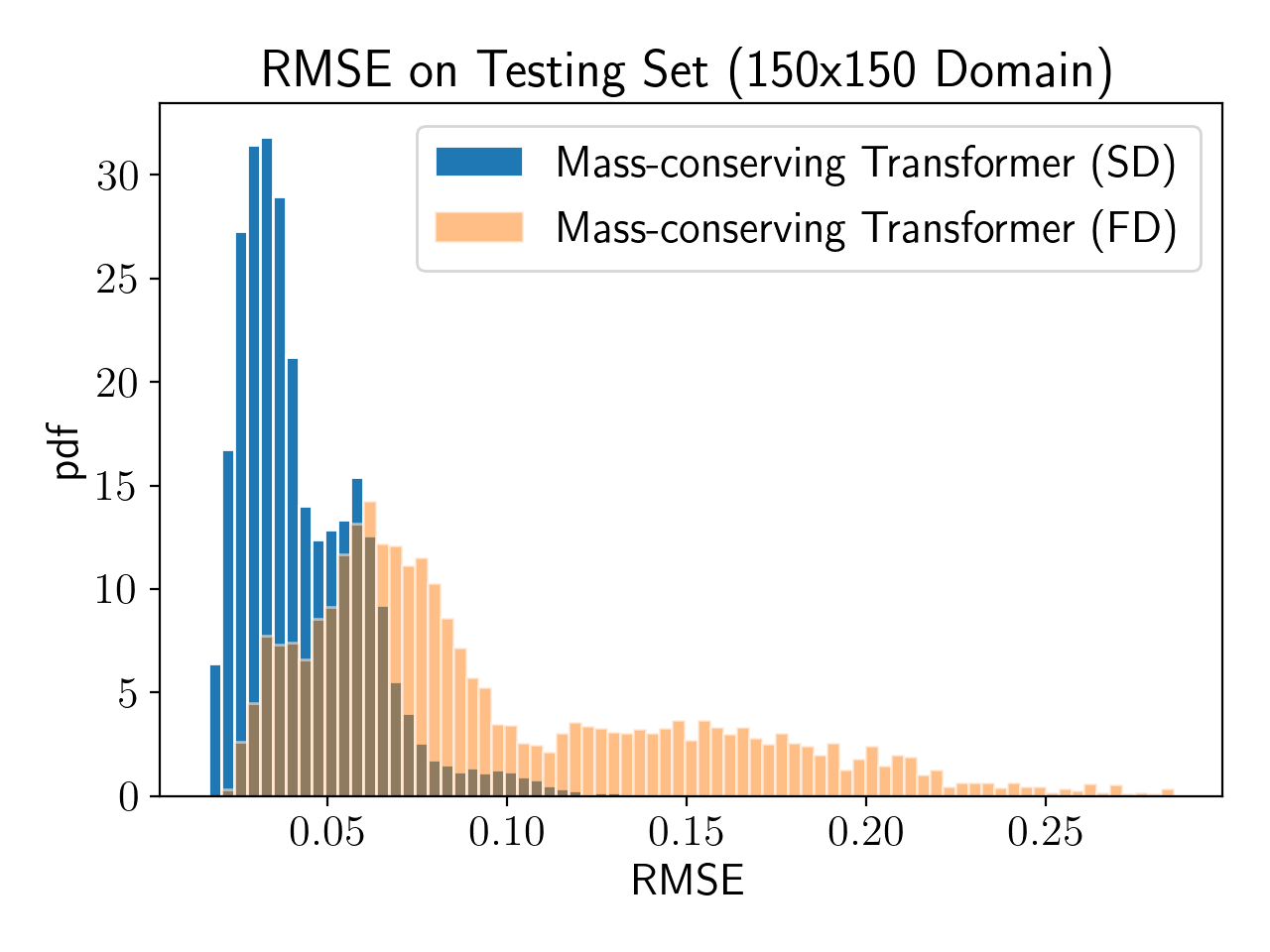}
  \includegraphics[trim=0 0 0 0, clip,
  width=.49\textwidth]{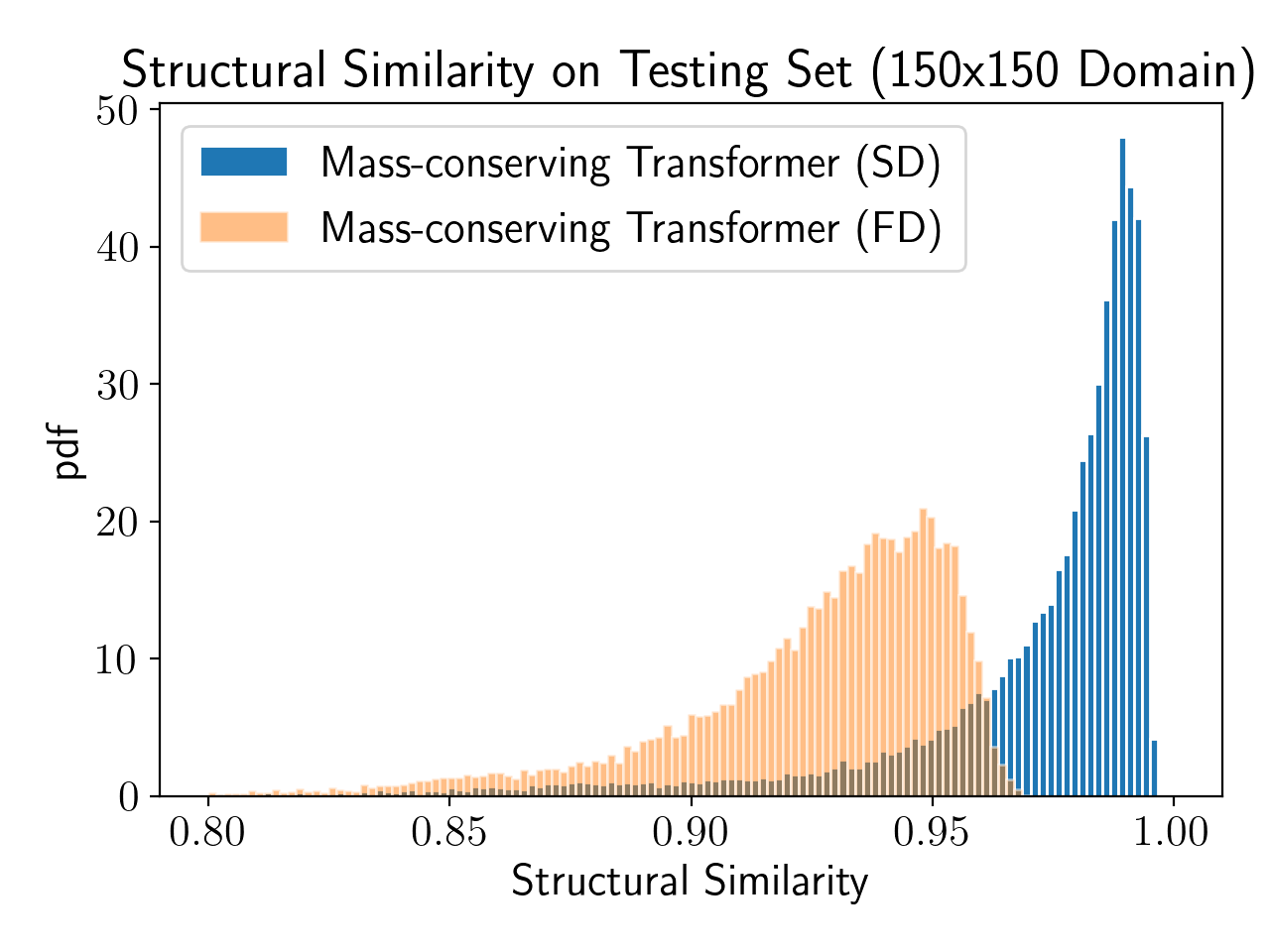} }
  \resizebox{\columnwidth}{!}{%
  \begin{tabular}{|c|c|c|c|c|c|c|}
  \hline
  \multirow{2}{*}{Architecture} 
  & \multicolumn{3}{c|}{RMSE (g/cc)}
  & \multicolumn{3}{c|}{Structural Similarity} \\
  \cline{2-7}
   & min & average & max & min & average & max\\
  \hline
Mass Conserving Transformer & 0.010 & 0.019 & 0.049 & 0.92 & 0.98 & 0.99 \\  
%Mass Conserving Transformer (FD) (440x440 Domain) & 0.021 & 0.040 & 0.104 & 0.92 & 0.98 & 0.99 \\% 0.35 & 0.67 & 1.73 & 0.92 & 0.98 & 0.99 \\
\hline
\remove{Mass Conserving Transformer (FD) (150x150 Domain) & 0.77 & 3.51 & 21.71 & 0.72 & 0.92 & 0.97 \\
\hline
Mass Conserving Transformer (SD) (150x150 Domain) & 0.59 & 1.51 & 5.16 & 0.76 & 0.97 & 1.00 \\}
\hline
  \end{tabular}%
  }
  \caption{Histogram of root-mean-squared errors (RMSE) (top left) and structural similarity (top right) between the
  density reconstruction and ground truth for the testing set
  for the Mass-conserving transformer trained on the 
  entire domain.
  Bottom: table of summary statistics for the above histograms.}
  \label{fig:histcomparisonfd}
\end{figure}

\begin{figure}[htb]
  \centering
  \includegraphics[trim=3cm .8cm 0cm 0, clip,
  width=.8\textwidth]{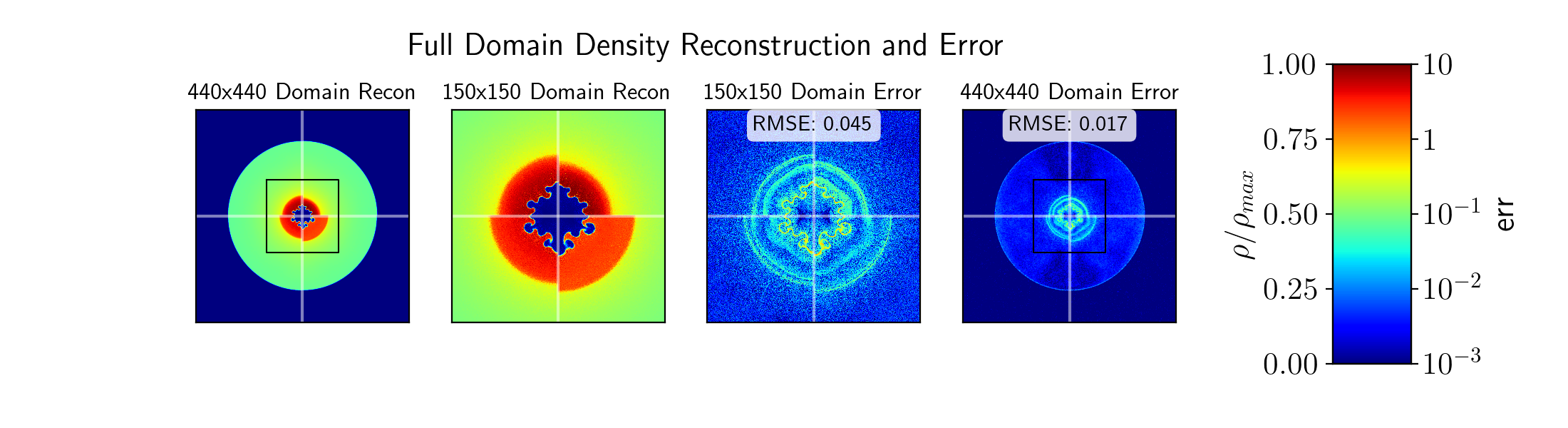}
  \caption{Example of full density reconstructions
  and their corresponding errors
  produced by the Mass-Conserving Transformer.
  The input shock features are polluted with 
  simulated noise from the radiograph-to-feature network.
  }
  \label{fig:fulldomainrecon}
\end{figure}

\begin{figure}[htb]
  \centering
  \includegraphics[trim=2.5cm 0cm 1.5cm 0, clip,
  width=.8\textwidth]{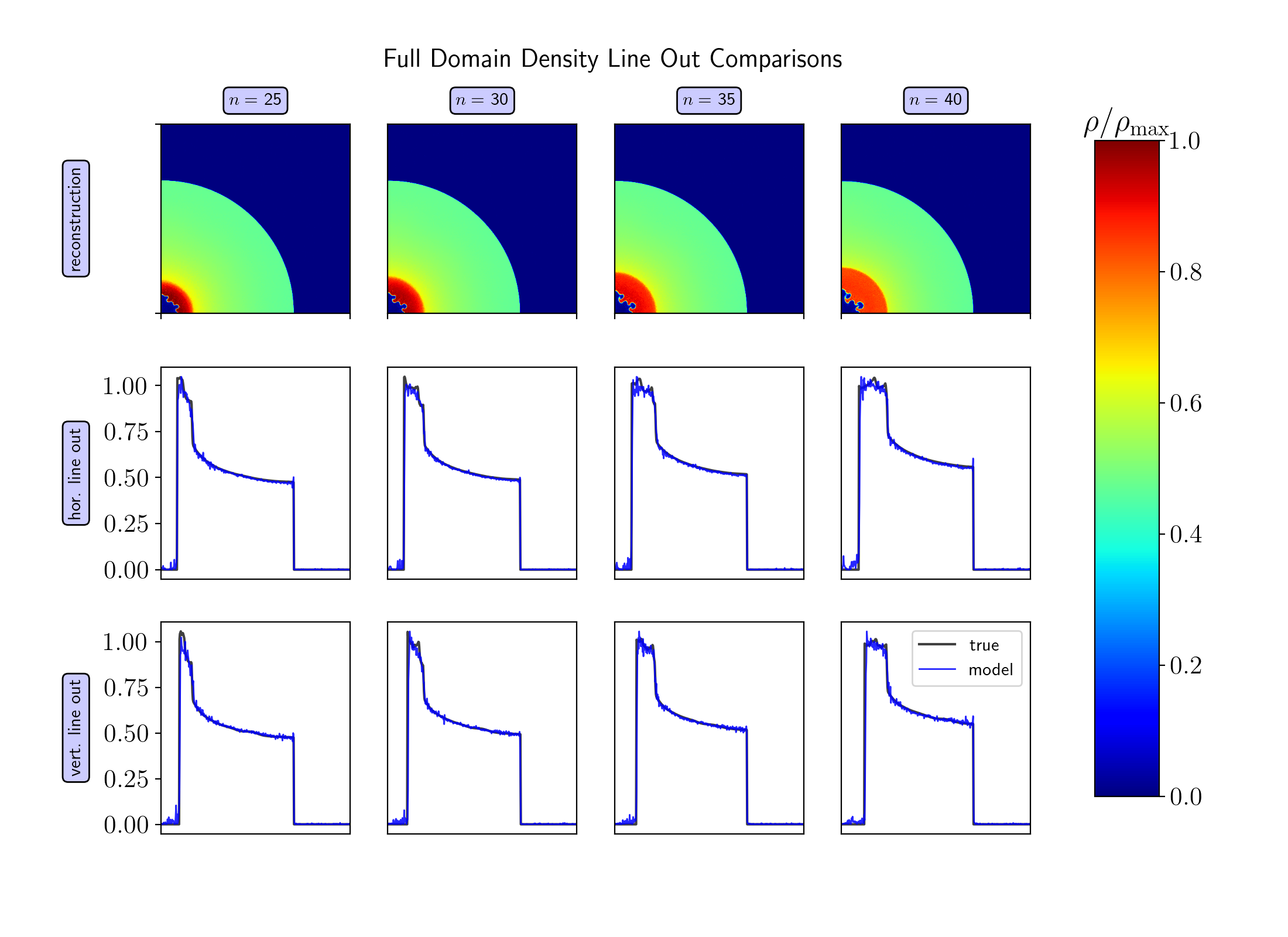}
  \caption{Plot of reconstruction and 
  horizontal and vertical density line plots
  produced by the 
  Mass-Conserving Transformer trained on the entire domain. The input shock features are polluted with 
  simulated noise from the radiograph-to-feature network.
  }
  \label{fig:lineprofile}
\end{figure}

While our previous results focused on reconstructing density in a smaller region around the RMI, we also demonstrate that our approach can be applied to reconstruct the density 
in the entirety of the domain using the Mass-Conserving Transformer.
A new model was trained using the
full $440 \times 440$ domain available in the data set
to produce a new set of architecture weights. 
The two plots in
Figure~\ref{fig:histcomparisonfd} show histograms of
RMSE and structural similarity evaluated on the 
testing set.
\remove{We compare the model trained on the full 440x440 domain 
(denoted with FD)
to the model trained on the smaller 150x150 domain
(denoted with SD)
by evaluating the RMSE and structural similarity 
on the $150 \times 150$ region.
These comparisons are shown in the 
bottom two histograms of 
Figure~\ref{fig:histcomparisonfd}.
Performance of the FD model is noticeably degraded 
in the smaller region around the RMI.}
We observe that the complexities of the gas metal 
interface are still reconstructed by the full 
domain model.
Figure~\ref{fig:lineprofile} 
shows an example of density reconstructions 
and corresponding horizontal and vertical line-outs
through the center and 
Figure~\ref{fig:fulldomainrecon}
shows a zoomed in view of the reconstruction
and its error.
In these examples, the RMI and shock location are still captured accurately.

\begin{figure}[htb]
  \centering
  \begin{subfigure}{\linewidth}
        \centering
  \includegraphics[trim=1.5cm .75cm 1.5cm 0cm, clip, width=.7\textwidth]{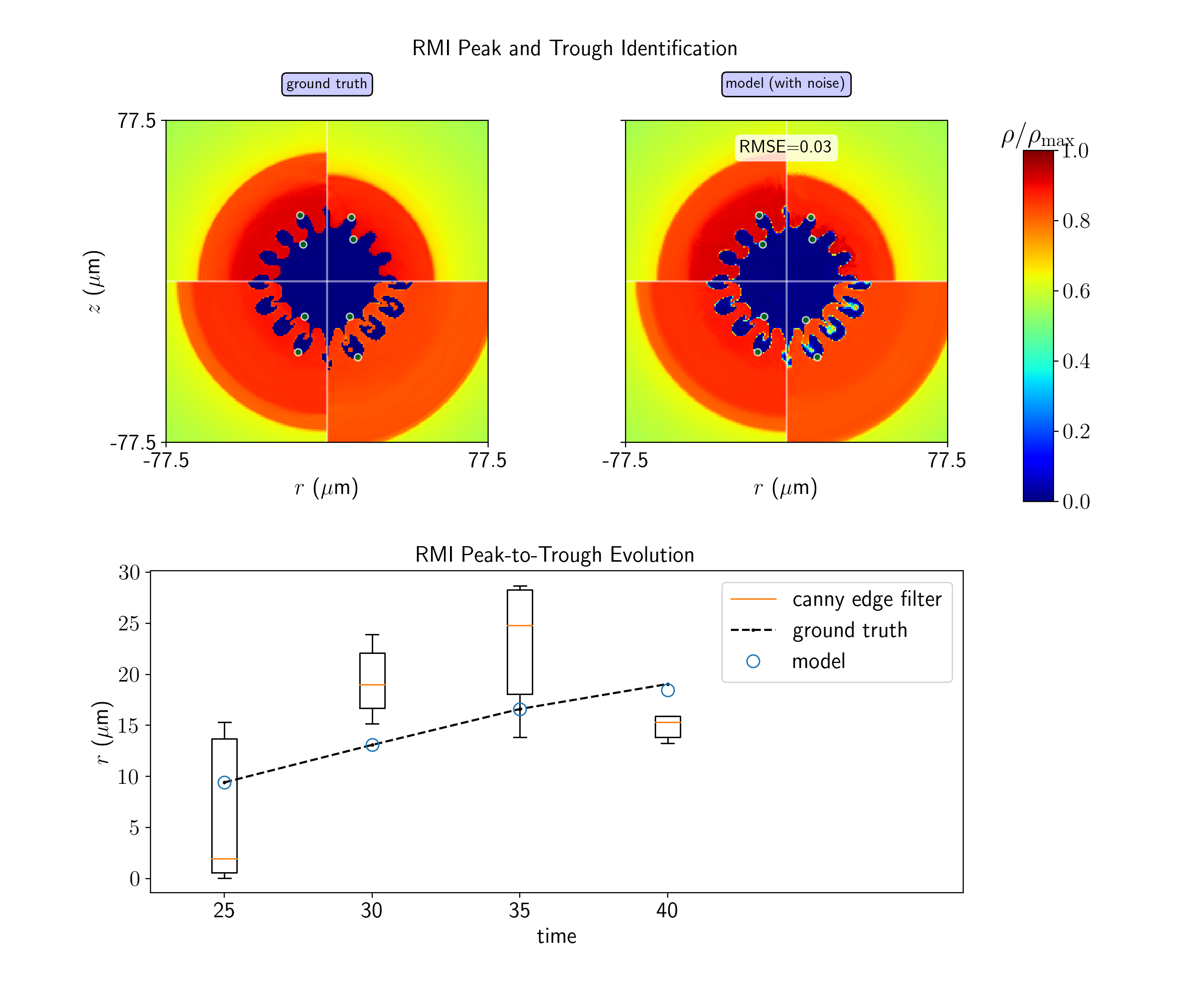}
  \caption{High frequency RMI mode example (profile 1).}
  \end{subfigure}
  \vfill
  \begin{subfigure}{\linewidth}
  \centering
  \includegraphics[trim=1.5cm .75cm 1.5cm 0cm, clip, width=.7\textwidth]{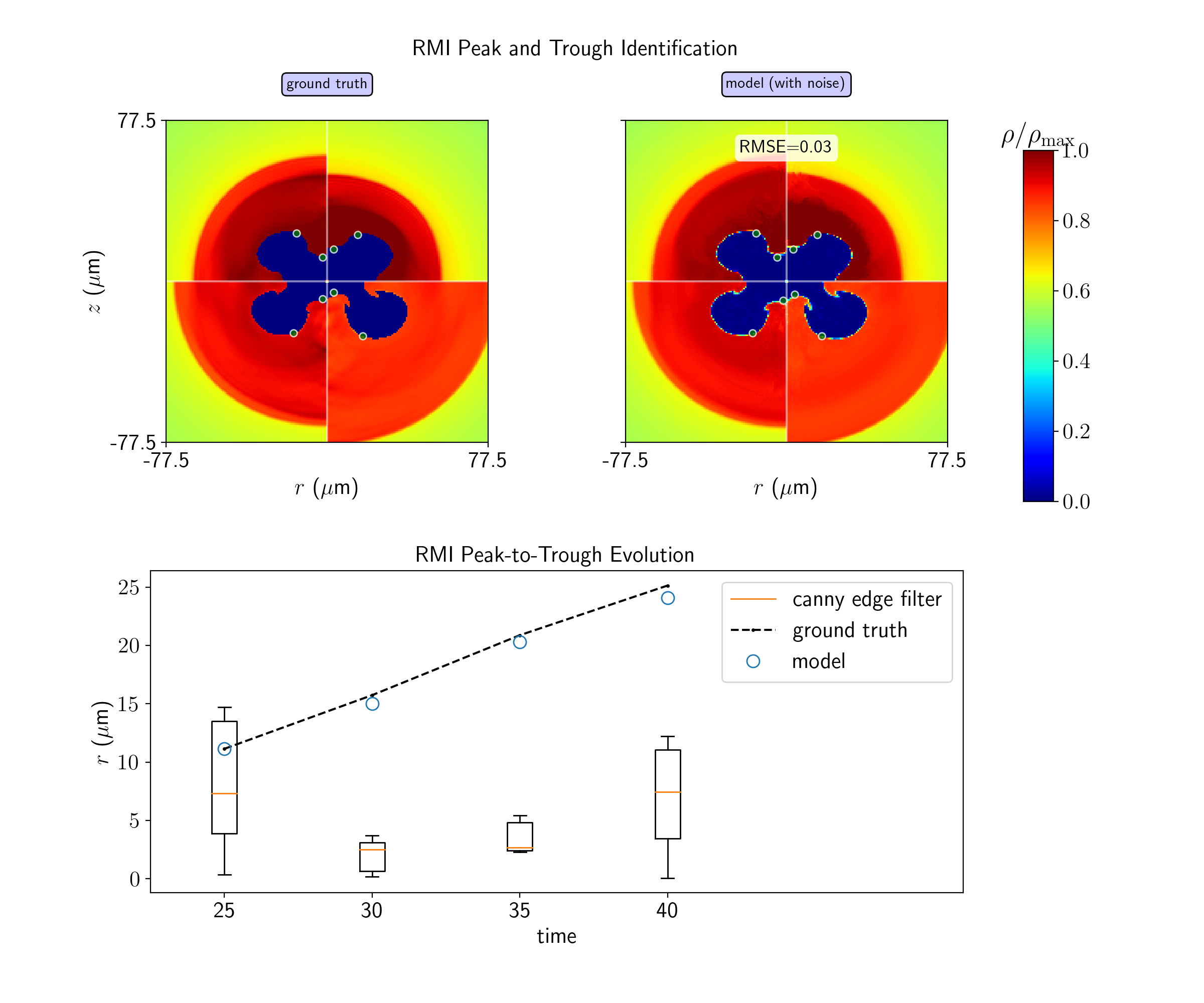}
  \caption{Low frequency RMI mode example (profile 9).}
  \end{subfigure}
  \caption{Top plots: density fields and identified
  peak and trough points of the RMI (green markers) 
  for the ground truth 
  and reconstructions using inputs polluted with 
  simulated noise from the radiograph-to-feature network.
  Bottom plots:
  evolution of the maximum RMI peak-to-trough radial distance 
  for the ground truth and reconstructions 
  corresponding to the above plots.
  }
  \label{fig:peaktrough}
\end{figure}

Next, we demonstrate that the Mass-Conserving Transformer can
accurately measure properties of the RMI
such as the growth of the peak-to-trough radial distance.
We chose a set of shock features,
applied a random perturbation generated from the noise model of the radiograph-to-features network,
and performed a density reconstruction. 
First we identified pixels that are contained in the 
gas domain, which is contained in a region 
near $(r,z)=(0, 0)$, which corresponds to a region
of significantly lower density compared to that of the surrounding 
metal.
A discontinuity separates the interface between the gas 
and metal.
On both sides of the discontinuity front, the densities
are approximately constant. 
In the reconstruction, there is a small region consisting of 
less than a few pixels where the density transitions smoothly
between the nominal fluid density and the nominal solid density.
Therefore, we attempt to split the difference.
We find the maximum density of the metal, $\rho_{\rm max}$
and take the fluid domain to be the region where the 
density is less than $\frac{1}{2} \rho_{\rm max}$.
From this, we identify the pixel locations on 
the edge of the domain with the furthest and
closest distances to the origin.
Figure~\ref{fig:peaktrough} shows the result of this identification 
algorithm applied to both the reconstruction and the ground truth for two examples of RMI profiles, 
representing high and low frequency modes.
Additionally, we apply a Canny edge filter
for various choices of the smoothing and threshold 
parameters to the 
noisy radiograph to identify the peak and trough of 
the RMI.
Figure~\ref{fig:peaktrough} shows a line plot comparing 
the peak-to-trough evolution between the ground truth,
the reconstruction, and Canny edge filter.
The density reconstruction approach is successful 
in accurately identifying the peak and trough locations
and consequently their radial distance.
The Canny edge filter however is unable to obtain an
acceptable level of accuracy and suffers from large
variations due to choices in meta-parameters. 
Figure~\ref{fig:sampledensitysingle} 
shows the resulting edges that are detected for the
high frequency profile (profile 1).
This demonstrates that our density reconstruction
algorithm is a viable approach 
for making growth-rate estimates in the evolution of 
a spherically-symmetric RMI.  We cannot overemphasize the importance of this discovery.  That is, this method of reconstruction may allow the RMI growth rates to be experimentally verified in a spherically convergent geometry for the first time.

\remove{
\begin{figure}[htb]
  \centering
\includegraphics[trim=0 0 0 0, clip, width=.45\textwidth]{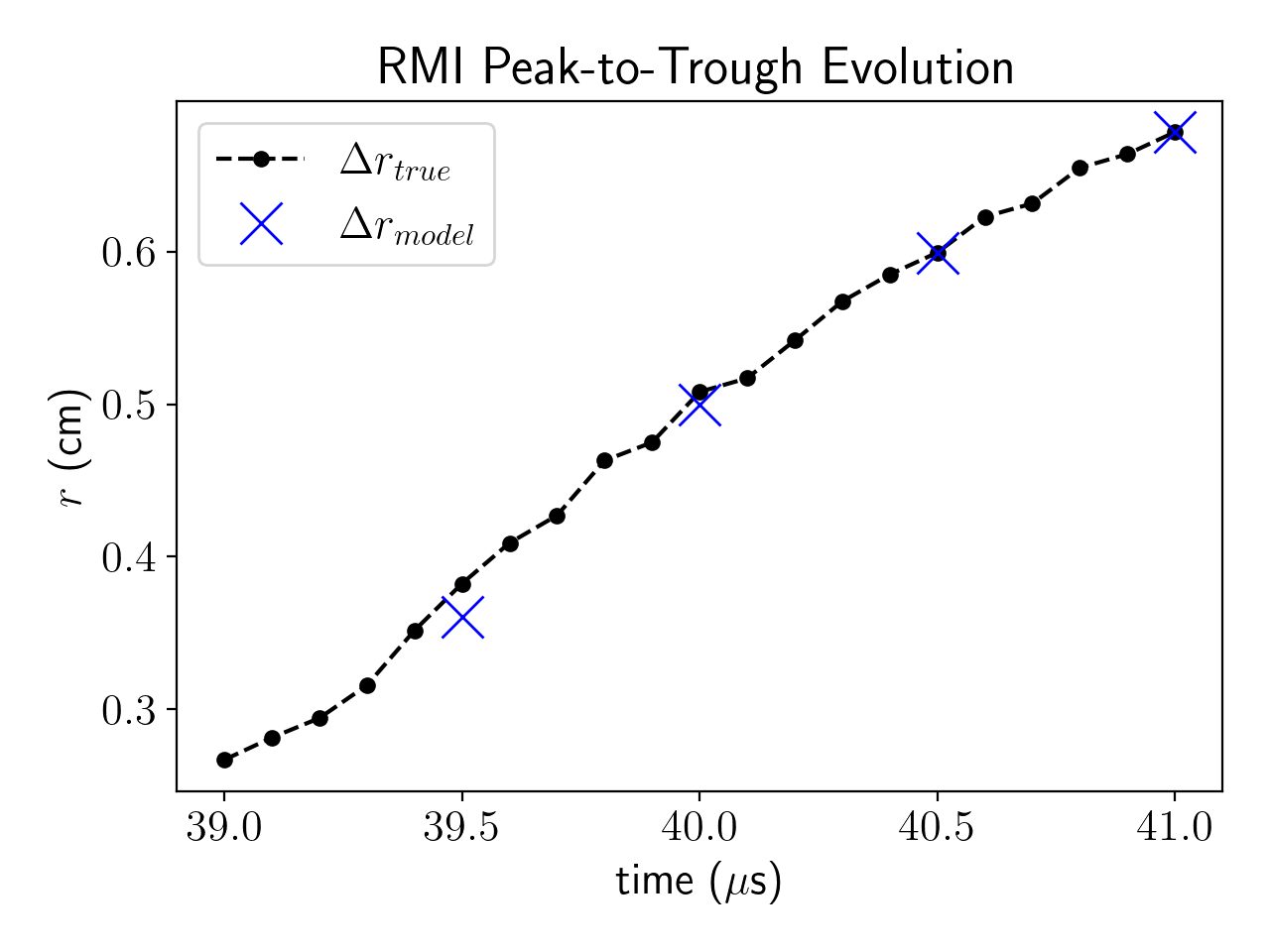}
  \caption{Comparison between the radius value for the trough and peak of the RMI for the density reconstruction and for the ground truth where the input shock features are polluted with 
  simulated noise from the radiograph-to-feature network.}
  \label{fig:peaktroughline}
\end{figure}  
}

\section{Conclusions}
\label{sec:Conclusions}

 This work presents a new density reconstruction approach
 that uses a trained attention-based transformer network to enable accurate density reconstructions
 from a series of noisy radiographic images.
 The key component of this network is the transformer encoder
 that acts on a sequence of features extracted from noisy radiographs.
 This encoder includes numerous self-attention layers
 that act to learn temporal dependencies in the input sequences and increase the expressiveness of the model. 

The two architectures had comparable performance,
with the Mass-conserving Transformer slightly outperforming the ShockDecoderViT.
Furthermore, the shock harmonic features provided sufficient constraints
to limit the variability in the density predictions,
with %indicated 
variations that are insignificant compared to the mean reconstruction errors.  

Our examinations also indicated the benefit of the dynamics
relative to the traditional approach of reconstruction of single images.
Finally, we demonstrated the ability to use our sparse features
to accurately reconstruct the fine features of the image,
enabling, for the first time, the determination of the RMI growth rates.
This work opens the potential to experimentally capture RMI growth rates in spherical geometries.  

%Our results demonstrate that this architecture can reconstruct density fields robustly and accurately, even in the presence of blur, scatter, and noise. In future studies, we plan to introduce further hydrodynamic structure in the model to improve its generalizability to inputs with different choices of hydrodynamic simulation parameters.

\appendix

\section{Cosine Coefficients for Inner Surface Perturbation Profile}

The coefficients of the cosine harmonic series of
the initial inner surface perturbation profile is scaled according to
$F_i = R_{\rm in} \bar{F}_i / 8,$ for
$i=0, \dots, 8$,
where $\bar{F}_0 = 8$, $\bar{F}_5=\bar{F}_7=0$,
and the rest of the coefficients are provided by Table~\ref{tab:initialcoeffs}.

\begin{table}
\begin{center}
\begin{tabular}{|c|c|c|c|c|c|c|c|c|}
\hline
profile& $F_1$ & $F_2$ & $F_3$ & $F_4$ & $F_6$ & $F_8$ \\
\hline
1 & 0 & 0 & 0 & 0 & 0 & 0.08\\
2 & 0 & 0 & 0 & 0.08 & 0 & 0\\
3 & 0 & 0.08 & 0 & 0 & 0 & 0\\
4 & 0 & 0 & 0 & 0 & 0 & 0.075\\
5 & 0 & 0 & 0 & 0.075 & 0 & 0\\
6 & 0 & 0.075 & 0 & 0 & 0 & 0\\
7 & 0 & 0.0075 & 0 & 0 & 0.0025 & 0.065\\
8 & 0.0075 & 0 & 0.0025 & 0.065 & 0 & 0\\
9 & 0.005 & 0.0657 & 0 & 0 & 0 & 0\\
10 & 0 & 0 & 0 & 0 & 0 & 0.06\\
11 & 0 & 0 & 0 & 0.06 & 0 & 0\\
12 & 0 & 0.06 & 0 & 0 & 0 & 0\\
13 & 0 & 0 & 0 & 0 & 0 & 0.055\\
14 & 0 & 0 & 0 & 0.055 & 0 & 0\\
15 & 0 & 0.055 & 0 & 0 & 0 & 0\\
16 & 0 & 0.0075 & 0 & 0 & 0.0025 & 0.045\\
17 & 0.0075 & 0 & 0.0025 & 0.045 & 0 & 0\\
18 & 0.0051 & 0.0457 & 0 & 0 & 0 & 0\\
19 & 0 & 0 & 0 & 0.04 & 0 & 0\\
20 & 0 & 0.04 & 0 & 0 & 0 & 0\\
\hline
\end{tabular}
\end{center}
\caption{Scaled cosine coefficients for the initial shell profile used for
  each profile. 
  }
\label{tab:initialcoeffs}
\end{table}

\begin{backmatter}
\bmsection{Funding}
This work was supported by the U.S. Department of Energy through the Los Alamos National Laboratory (LANL)
and the Laboratory Directed Research and Development program of LANL.

\bmsection{Acknowledgments}
The authors thank Jeff Fessler for helpful comments on the draft and Oleg Korobkin for extracting features for the data set. % :)

\bmsection{Disclosures}
The authors declare no conflicts of interest.

\bmsection{Data Availability Statement}
Data underlying the results presented in this paper are not publicly available at this time but may
be obtained from the authors upon reasonable request.

\end{backmatter}

\bibliography{bib, refs_mikepaper, sample}

\end{document}